\documentclass[journal]{IEEEtran}

\ifCLASSINFOpdf
\else
\fi

\usepackage{microtype}
\usepackage{amsmath,amssymb,amsfonts}
\usepackage{algorithmic}
\usepackage{graphicx}
\usepackage{textcomp}
\usepackage{bm}
\usepackage{subcaption}
\usepackage{pdflscape}
\usepackage{adjustbox}
\usepackage{arydshln}
\usepackage{array,multirow}
\usepackage{url}
\usepackage{booktabs}
\graphicspath{{Figures/}}

\newcommand{\footnotelabeled}[2]{%
	\addtocounter{footnote}{1}%
	\footnotetext[\thefootnote]{\label{#1}#2}}
\newcommand{\footnoteref}[1]{$^{\ref{#1}}$}

\begin{document}

\title{An Online Multi-Index Approach to Human Ergonomics Assessment in the Workplace}

\author{Marta Lorenzini, \IEEEmembership{Member, IEEE}, Wansoo Kim*, \IEEEmembership{Member, IEEE}, and Arash Ajoudani, \IEEEmembership{Member, IEEE}
\thanks{Manuscript received March 18, 2021.}

\thanks{This work was supported in part by the ERC-StG Ergo-Lean (Grant Agreement No.850932), in part by the European Union’s Horizon 2020 research and innovation programme under Grant Agreement No. 871237 (SOPHIA).}
\thanks{*Corresponding author.}
\thanks{All the authors are with the Human-Robot Interfaces and Physical Interaction Lab, Italian Institute of Technology, Genoa, Italy (e-mail: marta.lorenzini@iit.it). }
\thanks{W. Kim is also with Hanyang University, South Korea (e-mail: wansookim@hanyang.ac.kr). }}

\markboth{IEEE Transactions on Human-Machine Systems,~Vol.~XX, No.~X, MM~2021}%
{Lorenzini \MakeLowercase{\textit{et al.}}: An Online Multi-Index Approach to Human Ergonomics Assessment in the Workplace}

\maketitle

\begin{abstract}
Work-related musculoskeletal disorders (WMSDs) remain one of the major occupational safety and health problems in the European Union nowadays. Thus, continuous tracking of workers' exposure to the factors that may contribute to their development is paramount.
This paper introduces an online approach to monitor kinematic and dynamic quantities on the workers, providing on the spot an estimate of the physical load required in their daily jobs. A set of ergonomic indexes is defined to account for multiple potential contributors to WMSDs, also giving importance to the subject-specific requirements of the workers. To evaluate the proposed framework, a thorough experimental analysis was conducted on twelve human subjects considering tasks that represent typical working activities in the manufacturing sector. 
For each task, the ergonomic indexes that better explain the underlying physical load were identified, following a statistical analysis, and supported by the outcome of a surface electromyography (sEMG) analysis. A comparison was also made with a well-recognised and standard tool to evaluate human ergonomics in the workplace, to highlight the benefits introduced by the proposed framework. Results demonstrate the high potential of the proposed framework in identifying the physical risk factors, and therefore to adopt preventive measures. Another equally important contribution of this study is the creation of a comprehensive database on human kinodynamic\footnote{The term ``kinodynamic'' measurements is used, in this work, to indicate the variables associated with both human kinematics (i.e. positions, velocities, and accelerations) and dynamics (i.e. quantities related to interaction forces).} measurements, which hosts multiple sensory data of healthy subjects performing typical industrial tasks.
\end{abstract}


\IEEEpeerreviewmaketitle

 \section{Introduction}
\label{sec:introduction}
\IEEEPARstart{I}{n} 21\textsuperscript{th} century, the production lines of a large proportion of the manufacturing enterprises worldwide still rely on processing and assembly activities that are performed manually by human operators \cite{frohm2008levels}.
As a result, an alarming worsening of the workers' health conditions in the last decade has been reported \cite{bortolini2018automatic}. 
Indeed, roughly three out of every five workers in the European Union (EU) report on a daily basis tiring or even painful conditions while performing their regular tasks, which frequently result in critical chronic work-related musculoskeletal disorders (WMSDs) \cite{EUOSHA2019}.
WMSDs are not just conditions of older age but also affect workers across the life-course, in all sectors and occupations, with an extremely heavy charge to companies and healthcare systems \cite{james2018global,punnett2004work}. 
Hence, the careful monitoring of workers' exposure to the factors that may contribute to their development is a key requirement in the manufacturing sector, aiming to lay the foundation of risk prevention and reduction programs. 

\begin{figure}[t]
	\centering
	\vspace{-3 mm}
	\includegraphics[trim={14.5cm 6cm 5cm 2cm},clip,width=0.45\textwidth]{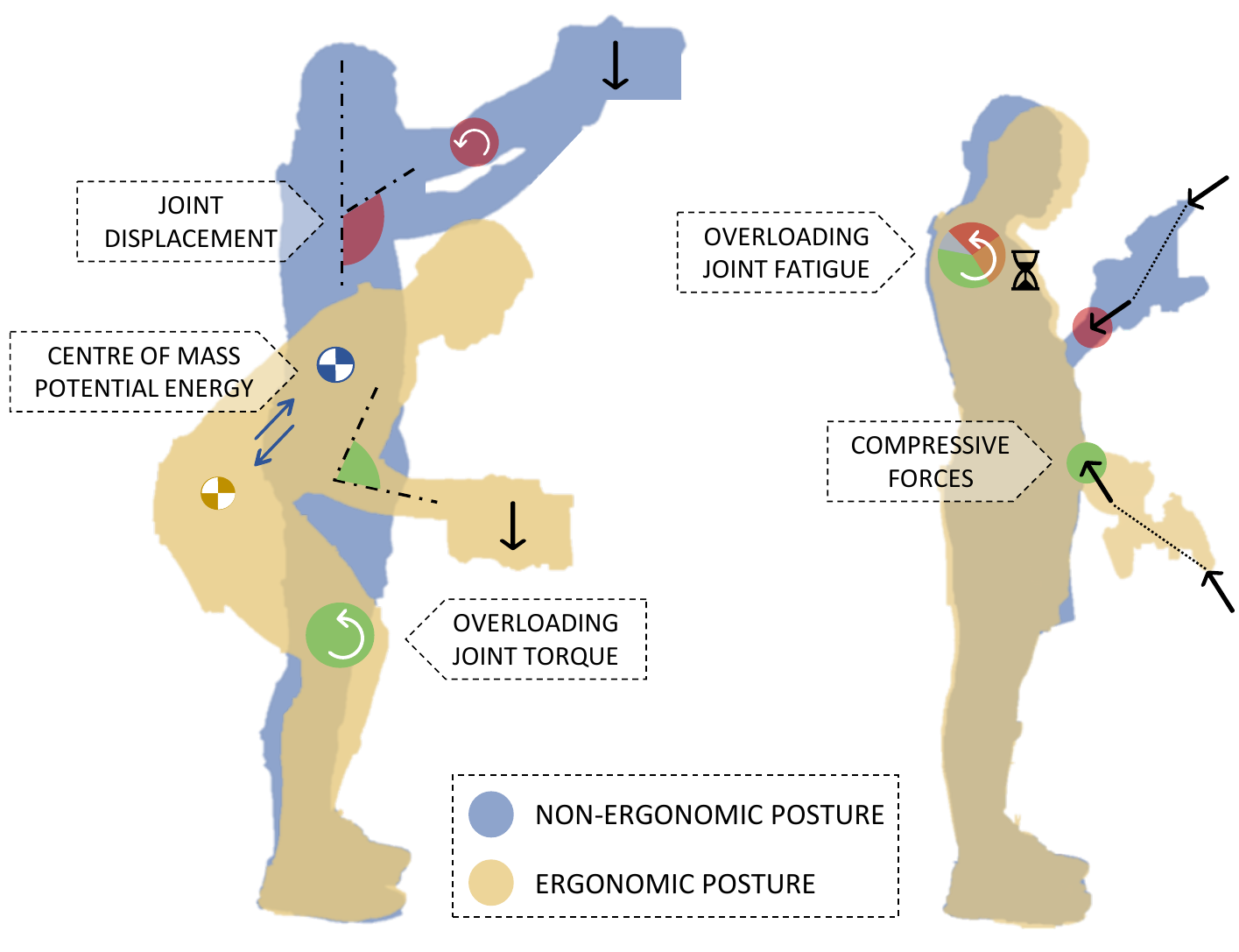}
	\label{fig:scenario001}
    \caption{An online multi-index approach is proposed to monitor both kinematic and dynamic quantities on the human whole body. Setting proper thresholds for such indexes, the associated ergonomic risk can be categorised and more convenient body configurations and loading conditions can be identified.}
	\label{fig:concept}
	\vspace{-5 mm} 
\end{figure}

To this end, a considerable number of methods and approaches have been proposed by researchers over the years \cite{li1999current,david2005ergonomic}.
Concerning the industrial scenario, the great majority of the tools currently adopted to evaluate workers' ergonomics relies on the so-called ``pen-and-paper'' observational techniques. Most of these analyse a particular aspects or a specific activity 
(NIOSH: carrying and lifting, Snook and Ciriello: pushing and pulling, OCRA: low loads at high frequency, OWAS, RULA \& REBA: posture and movements \cite{david2005ergonomic}), however such a high specificity severely limits their application field. Conversely, the more recently developed Ergonomic Assessment Worksheet (EAWS) \cite{schaub2013european} provides a unique ergonomic score considering workers' activities in a more comprehensive view. 
However, all these methods must be performed as an off-line procedure and mainly focus on the kinematics of the activities (i.e. body postures and movements) while the dynamic aspects are deemed to a limited extent. 
On the other hand, to consider also the moments and forces developed within the body due to external interactions, many studies were conducted wherein direct measurements collected on the subjects are integrated with models of the human body. Most of the proposed approaches adopt detailed biomechanical models of the human musculoskeletal structure to estimate dynamic states such as joint reactions (forces and torques) by using inverse dynamics and then optimisation techniques to compute the muscle tensions \cite{delp2007opensim,damsgaard2006analysis}. Alternatively, using electromyography (EMG) signals the muscle activity can be measured directly, and then empirical models can be exploited to convert such activity into muscle tensions \cite{murai2010musculoskeletal,sartori2012emg}. Nevertheless, all the models underlying these techniques require the estimation of a large number of parameters, or otherwise, they can be obtained by means of anthropometric standards \cite{winter1995human} thus the achieved estimated quantities are not subject-specific. Moreover, the sensor system employed (e.g. EMG, marker-based optical motion-capture) are often cumbersome and the procedures involved are computationally expensive, affecting the online capabilities of most of them. 

In the brand-new industrial background, the accurate assessment of a workers' ergonomics should be performed via noninvasive sensing technologies that require short preparation time as well as advanced yet rapid probing techniques that can provide instant data on human physical load \cite{ajoudani2020smart}. In light of this, many researchers have recently focused their effort on developing online and practical strategies to account for human physical ergonomic risk factors by means of robotics-inspired indexes \cite{maurice2017human,rapettipartner} but also low-dimensional space models \cite{marin2018optimizing} or exploiting vision-based algorithms \cite{parsa2019toward,liu2016tracking}. 

In the same line, the foremost objective of this paper is to introduce an online human ergonomics monitoring framework that can provide on the spot, just as the workers are carrying out their activities, an estimate of the required physical load. An exhaustive set of indexes is defined that account for multiple ergonomic risk factors (e.g. awkward postures, mechanical overloading of the body, repetition frequency, etc.) and thus allow the evaluation of different manual activities typically conducted in the manufacturing sector.
Besides considering kinematic variables such as joint angles, velocities, and accelerations, a great emphasis is placed on the dynamic aspects of the tasks (i.e. forces or torques acting on and inside the body). Indeed, some indicators that were investigated and developed in previous studies \cite{kim2017real,lorenzini2019new,fortini2020compressive} such as overloading torque, fatigue, and power, the human centre of mass (CoM) potential energy, and compressive forces are integrated into the framework to establish a comprehensive method for the assessment of humans' ergonomics in the workplace (see Figure \ref{fig:concept}). The proposed framework is conceived to work with wearable and easy-to-use sensor systems, which allow the users to move freely and can be used for long periods, meeting the requirements of real factories. Furthermore, to address workers' individual demands, the estimation of such ergonomic indexes is based on a subject-specific and fast re-identifiable human model.

To evaluate the ergonomics monitoring framework introduced in this study, an experimental analysis was conducted on twelve human subjects. Participants were required to perform three different tasks that represent typical working activities in the manufacturing sector and, additionally, are associated with different potential risk factors to the development of WMSDs. 
The proposed set of indexes was computed and then, their outcome was investigated to establish their link with the most common ergonomic risk factors. It should be underlined that, as previously said, the proposed indexes are conceived to perform an online evaluation, as demonstrated in our previous works \cite{kim2017real,lorenzini2019new,fortini2020compressive}. However, in this study, they are computed a posteriori for the sake of a thorough analysis. A statistical method was also employed to verify, for each ergonomic index, the significance of the differences among the different experimental conditions for each task. Indeed, in this work, we were mainly interested in the analysis of the indexes by varying the physical load (i.e. experimental conditions) to identify their capability to address the workload associated with a specific activity. Thus, we focus more on the comparison among different conditions than on the indexes' absolute values.
In addition, to provide a benchmark of the effective physical effort required for the performed activities, muscle activity was monitored through a sEMG system over the entire duration of the experiments. Since the proposed indexes are expressed on the joint level and cannot be directly compared to sEMG signals that are on the muscle level, the overall trend of the muscle activity was compared with the trend of the developed metrics. Finally, a comparison was made with the ergonomic risk scores resulting from the EAWS \cite{schaub2013european} inasmuch as they are a well-recognised method to evaluate workers' physical exposure to WMSDs. 

Due to the growing interest in occupational ergonomics on one hand, and the effort needed to create and organise a database, on the other, numerous human motion datasets on industry-oriented activities were recently produced and made available to researchers and ergonomics practitioners. 
To cite a few, in \cite{mandery2015kit,maurice2019human} comprehensive whole-body motion capture datasets that include trials of typical working tasks (e.g. manipulating a screw, driver or hammer, lifting loads) were proposed. In view of this, another contribution of this study is the creation of a comprehensive database that contains measurements collected with multiple sensors systems on healthy human subjects performing the aforementioned working activities. Both kinematic, dynamic and sEMG quantities are included, as well as videos recorded to facilitate the management of the data by future users.

The rest of the paper is organized as follows. In Section \ref{sec:overview}, an overview of the proposed human ergonomics monitoring framework is provided.
Section \ref{sec:experiments} illustrates the experimental analysis conducted to evaluate the framework. Section \ref{sec:database} describes the developed human measurements database on factory-like activities. In Section \ref{sec:conclusion}, the contribution and future improvements of this study are highlighted.

\section{Method Overview}
\label{sec:overview}

In this Section, the human body model adopted in this study is first presented. Then, a set of kinodynamic indexes to monitor online the humans' physical load during daily working activities is introduced.

\subsection{Human Modeling Bases}

The framework introduced in this paper has been conceived in view of its application and usage in a real industrial environment. Hence, it aims at a simplified approach that focuses on fast identifiability of the models and online capabilities of the methods, respectively.
Based on that, the human model adopted in this study is a reduced-complexity representation of the human musculoskeletal structure. Specifically, it is a floating-based sequence of rigid links interconnected by revolute joints and limited to the sagittal plane. 
The pelvis frame is set as the human base frame $\Sigma_{0}$ and it is attached to the inertial frame $\Sigma_{W}$ through six virtual degrees of freedom (DoFs). 
$\boldsymbol{q}=[\boldsymbol{x}_0^T \enspace \boldsymbol{\theta}_0^T \enspace \boldsymbol{q}_h]$ refers to the generalized coordinate of the system. $\boldsymbol{x_0} \in \mathbb{R}^3$ and $\boldsymbol{\theta}_0 \in \mathbb{R}^3$ are the position and the orientation of the human base frame $\Sigma_{0}$. The rigid links are articulated through $n_{j}$ revolute joints whose angular position is denoted by $\boldsymbol{q}_h=[q_1 \enspace \ldots \enspace q_{n_{j}}]^T\in \mathbb{R}^{n_{j}}$. 
To obtain the human body segment inertial parameters (BSIPs), a reduced-complexity approach called statically equivalent serial chain (SESC) technique \cite{cotton2009estimation} is employed in this study.
The SESC parameters can be employed to achieve a subject-specific estimation of the whole-body CoM.
As a result, the whole-body centre of pressure (CoP) estimation, which is required in the computation of some of the ergonomic indexes that will be addressed in the next paragraph, can be accomplished. 
In static conditions, the CoP can be computed by simply projecting the whole-body CoM onto the $x$-$y$ plane. Conversely, when considering dynamic conditions, the position of CoP w.r.t. the CoM can be obtained by using the differences between the angular momentum variations and the acceleration about the CoM 
\cite{popovic2005ground}. 
A key strength of this method is that, due to its simplified approach, it can be applied online \cite{kim2017anticipatory}.

\subsection{Set of Ergonomic Indexes}

By integrating the data about human motion and interaction forces, which can be collected through suitable sensor systems, within a model of the human body, it is possible to estimate several kinematic and dynamic quantities (e.g. joint position/velocity/acceleration, joint overloading torque/fatigue) that can be linked to ergonomic targets such as mechanical overloading of the musculoskeletal structure and body posture. Hence, the proposed human ergonomic monitoring framework is defined as a set of indexes, which are expressed as $\bm{\omega}_{h}(\mathbf{q}) \in \mathbb{R}^{n_{j}}$, that account for the physical exposure of the workers to different risk factors to the development of WMSDs. Such indexes, with the corresponding equation, scope, and the ergonomic target that each one of them seeks to address are listed in Table \ref{tab:indexes}. In general, $\mathbf{X}^{\text{max}}$ and $\mathbf{X}^{\text{min}}$ denote the vectors including the maximum and minimum values of the variable $X$, respectively. $|\mathbf{X}|$ denotes instead the vector including the absolute values of the variable $X$.

\begin{table}[h]
\centering
\caption{The proposed set of ergonomic indexes to assess human kinodynamic quantities accounting for multiple ergonomic risk factors. The corresponding scope (kinematics or dynamics), equation and ergonomic target are included for each one of them.}
\label{tab:indexes}
\begin{adjustbox}{max width=0.49\textwidth}
\begin{tabular}{cccc}
\toprule \\
\Large{SCOPE} & \Large{INDEX} & \Large{EQUATION} & \Large{RISK FACTOR}\\ \\
\midrule \\
\multirow{3}{*}{\rotatebox[origin=c]{90}{\Large{Kinematics}}} & \textbf{\large{Joint Displacement}} &  {\Large{ ${\bm{\omega}}_{1}(\mathbf{q}) =  \frac{|\mathbf{q}_h| }{\mathbf{q}_h^{\text{max}} - \mathbf{q}_h^{\text{min}}}$} } & \normalsize{Awkward body postures} \\ \\
& \textbf{\large{Joint Velocity}} & {\Large ${\bm{\omega}}_{2}(\mathbf{q}) =  \frac{\dot{\mathbf{q}_h}}{\dot{\mathbf{q}_h}^{\text{max}}}$ } & \normalsize{Abrupt and sudden efforts} \\ \\
& \textbf{\large{Joint Acceleration}} & {\Large${\bm{\omega}}_{3}(\mathbf{q}) = \frac{\ddot{\mathbf{q}_h}}{\ddot{\mathbf{q}_h}^{\text{max}}}$ } & \normalsize{\begin{tabular}[c]{@{}c@{}}High-intensity forces \\ due to inertia\end{tabular}} \\ \\ \midrule 
 & \begin{tabular}[c]{@{}c@{}}\textbf{\normalsize{Overloading}} \\ \textbf{\large{Joint Torque}}\end{tabular} & {\Large${\bm{\omega}}_{4}(\mathbf{q}) =  \frac{\Delta{\bm{\tau}}}{\Delta{\bm{\tau}}^{\text{max}}} $} & \normalsize{\begin{tabular}[c]{@{}c@{}}Mechanical overburden \\ of the musculoskeletal stucture\end{tabular}} \\ \\ \cmidrule{2-4}
 & \begin{tabular}[c]{@{}c@{}}\textbf{\large{Overloading}} \\ \textbf{\large{Joint Fatigue}}\end{tabular} & {\Large${\bm{\omega}}_{5}(\mathbf{q}) =  \frac{\bm{\tau}^{F}(t)}{\bm{\tau}^{F,\text{max}}(t)}$} & \normalsize{\begin{tabular}[c]{@{}c@{}}Repetitive and monotonous \\ movements \end{tabular}}\\ \\ \cmidrule{2-4}
\multirow{-3}{*}{\rotatebox[origin=c]{90}{\Large{Dynamics}}}& \begin{tabular}[c]{@{}c@{}}\textbf{\large{Overloading}} \\ \textbf{\large{Joint Power}}\end{tabular} & {\Large${\bm{\omega}}_{6}(\mathbf{q}) = \frac{\mathbf{P}}{\mathbf{P}^{\text{max}}} =  \frac{\dot{\mathbf{q}_h}\Delta{\bm{\tau}}}{\dot{\mathbf{q}_h}^{\text{max}}\Delta{\bm{\tau}}^{\text{max}}}$} & \normalsize{\begin{tabular}[c]{@{}c@{}}Mechanical overburden \\ of the musculoskeletal stucture\end{tabular}} \\ \\
 & \begin{tabular}[c]{@{}c@{}}\textbf{\large{CoM}} \\ \textbf{\large{Potential Energy}}\end{tabular} & {\Large${\bm{\omega}}_{7}= \frac{\Delta {E}_{P}}{ \Delta {E}_{P}^{\text{max}}} = \frac{\Delta C_{M}|z}{ \Delta C_{M}|z^{\text{max}}}$} & \normalsize{Awkward body postures} \\  \\ \cmidrule{2-4}
 & \textbf{\large{Compressive Forces}} & {\Large${\bm{\omega}}_{8}(\mathbf{q}) = \frac{\mathbf{f}_{C}}{\mathbf{f}_{C}^{\text{max}}} $} & \normalsize{\begin{tabular}[c]{@{}c@{}}Mechanical overburden \\ of the musculoskeletal stucture\end{tabular}} \\ \\ \bottomrule
\end{tabular}
\end{adjustbox}

\end{table}

It should be noted that the indexes listed in Table \ref{tab:indexes} are all normalised, i.e. divided by their corresponding maximum values, to concurrently analyse and compare their outcome. In addition, as indicated in the leftmost column, they can be divided into two categories: kinematics and dynamics, respectively. The explanation (with reference to the equations reported in Table \ref{tab:indexes}), as well as the corresponding rationale of each ergonomic index, will be addressed as follow.

\subsubsection{Kinematic Indexes}

The first category includes the variables related to the kinematics (i.e. motions without regard to forces that cause it) of the workers' activities.

The first index $\bm{\omega}_{1}(\mathbf{q})$ is the joint displacement, which is considered here as the distance from the mechanical joint limits. Given the joint angles vector $\mathbf{q}_h$ as defined in the previous paragraph, $\bm{\omega}_{1}(\mathbf{q})$ can be computed considering its absolute values $|\mathbf{q}_h|$ and the joint angles upper and lower boundaries, $\mathbf{q}_h^{\text{max}}$ and $\mathbf{q}_h^{\text{min}}$, respectively, which can be found in literature \cite{whitmore2012nasa}.
The aim of monitoring the joint displacement is to detect wherever the body configurations that workers adopt to perform their tasks are not ergonomic. In fact, certain human postures have proven to be a potential cause of musculoskeletal diseases \cite{lutmann2003protecting}. 
The latter are mostly related to specific sections of the human range of motion (RoM) that should be avoided (e.g. in proximity to the maximum limits). 

Next, the second index $\bm{\omega}_{2}(\mathbf{q})$ is the joint normalised angular velocity, where $\dot{\mathbf{q}}_h \in \mathbb{R}^{n_{j}}$ is the joint velocities vector and $\dot{\mathbf{q}_h}^{\text{max}}$ is the vector including the joints maximum velocities. The values of $\dot{\mathbf{q}_h}^{\text{max}}$ can be found in literature \cite{jessop2016maximum}.
Finally, the third index $\bm{\omega}_{3}(\mathbf{q})$ is the joint normalised angular acceleration, where $\ddot{\mathbf{q}}_h \in \mathbb{R}^{n_{j}}$ is the joint accelerations vector and $\ddot{\mathbf{q}}_h^{\text{max}}$ is the vector including the joints maximum accelerations. The values of $\ddot{\mathbf{q}}_h^{\text{max}}$ can be obtained experimentally, by asking the subjects to move dynamically to the maximum extent possible, exciting all the joints, and recording the maximum accelerations achieved.
As reported by some researchers, when assessing dynamically varying activities, the direct measurement of human motion - in terms, for example, of joint velocity and acceleration - can provide more relevant information than posture analysis for the ergonomic risk of the low back \cite{marras1995biomechanical} and upper limbs \cite{malchaire1997relationship}.
Joint velocities, in particular, are suggestive of eventual abrupt movements that may be frequently associated with sudden and acute efforts potentially leading to severe injuries. On the other hand, the forces induced in the human joints due to the inertia of the body masses are directly proportional to the accelerations (e.g. Newton's second law) and may lead to the overload of musculoskeletal tissues. This can happen, for example, when the workers perform actions that require high accelerations to be accomplished efficiently (e.g. hammering). 

\subsubsection{Dynamic Indexes}

The second category includes the variables related to the dynamics (i.e. motions that result from forces) of the workers' activities.

The fourth index arises from the ``overloading joint torque'' method that was originally proposed in \cite{kim2017real} to account online for the torque variations induced on the human main joints by an external heavy load. Its computation is based on the displacement of the CoP, computed from the difference between an estimated one and a measured one. The estimated CoP vector $\hat{\bm{C}}_P$ can be obtained as explained in the previous paragraph, by taking advantage of the SESC technique. On the other hand, the measured CoP vector ${\bm{C}}_{P}$ can be collected using an external sensor system. 
Details of the method can be found in \cite{kim2017real}.
Specifically, ${w}_{4}(\mathbf{q})$ is the joint normalised overloading torque, where $\Delta{\bm\tau} \in \mathbb{R}^{n_{j}}$ is the overloading joint torque vector and $\Delta{\bm\tau}^{\text{max}}$ is the vector including the maximum values of the human overloading joint torque, which can be estimated directly through experiments on the subjects.
Increasing torque profiles are applied on the selected body joints, one at a time, until the subjects start to feel discomfort. In such a specific instant, the resulting torque values are estimated (based on the applied force and the lever arm) and compared to the ones extracted from literature
\cite{anderson2007maximum}. 
If these values are comparable, the experimental ones are used as the maximum torque values. If the differences are significant, the ``safest'' choice, i.e. the smallest value for the maximum torque, is chosen. 
The overloading joint torque index $\bm{\omega}_{4}(\mathbf{q})$ can be employed to account for the mechanical overburden of musculoskeletal structures induced by the weight of a tool or object. It should be pointed out that this method takes into account only the vertical component of external forces thus it can be deployed in a certain class of industrial tasks, which nevertheless, are quite numerous (lifting/lowering, carrying, pick and place, etc.).

The fifth index is the ``overloading joint fatigue'' that was originally introduced in \cite{lorenzini2019new} as an extension of the previous index. 
Indeed, when dealing with a light payload (e.g. a lightweight tool), the overloading torques induced on the joints are low/moderate and the associated risk of injuries is not significant. On the other hand, 
the building up of their effect over a protracted period of time could become hazardous thus a whole-body fatigue model was developed to evaluate this aspect. 
Details of the method can be found in \cite{lorenzini2019new}. 
Specifically, ${\bm\omega}_{5}(\mathbf{q})$ is the joint normalised overloading fatigue, where $\bm{\tau}^{F}\in \mathbb{R}^{n_{j}}$ is the overloading joint fatigue vector and $\bm\tau^{F,\text{max}}$ is the vector including the maximum values of the human overloading joint fatigue. Since fatigue is strictly related to the subject’s physical capacity and feelings, the maximum endurance time (MET) concept presented in \cite{imbeau2006percentile} was employed to find experimentally the subject-specific parameters of the model (see \cite{lorenzini2019new}). 
The accumulation of local joint fatigue due to external load, which can be monitored by means of ${\bm\omega}_{5}(\mathbf{q})$, can result from repetitive and monotonous actions, which are among the most frequently cited contributors to WMSDs \cite{bernard1997musculoskeletal}. 

The sixth index ${\bm\omega}_{6}(\mathbf{q})$ is the ``overloading joint power'' that basically combines the second and the fourth indexes previously explained.
Specifically, ${\bm\omega}_{6}(\mathbf{q})$ is the joint normalised overloading power where $\mathbf{P}\in \mathbb{R}^{n_{j}}$ results from the overloading joint torque vector $\Delta{\bm{\tau}}$ multiplied by the joint velocity vector $\dot{\mathbf{q}}_h$, whereas $\mathbf{P}^{\text{max}}$ results from $\Delta{\bm\tau}^{\text{max}}$ multiplied by $\dot{\mathbf{q}_h}^{\text{max}}$ and includes the corresponding maximum values of $\mathbf{P}$.
Based on human joint power, several procedures can be found in the literature with the aim to estimate the physical effort needed during gait \cite{ren2007predictive} but also to detect potential risk in manual material handling \cite{gagnon1991muscular}. Overloading joint power is different from joint power in his classical conception since the overloading joint torques (which takes into account only the effect of an external load) and not the net joint torques are employed in the computation.
Nevertheless, it is considered worthwhile to investigate also ${\bm\omega}_{6}(\mathbf{q})$ as a possible indicator of the physical expenditure during occupational activities.

The seventh index ${\bm\omega}_{7}$ is the ``CoM potential energy'' that is developed based on the study presented in \cite{yang2004multi}, wherein the change in potential energy of human body masses was employed as a measure of performance within an optimisation-based approach to model human motion. The potential energy for the CoM of the human body can be obtained by multiplying the height of the CoM, thus the $z$-coordinate, by the force of gravity and the human body mass. However, such a quantity is not evaluated directly but rather its variation between different body configurations is considered. Accordingly, two potential energies are defined: ${E}_{P}^{0} = mg^{T}C^{0}_{M}|z$ that is associated with a neutral and natural body configuration and ${E}_{P} = mg^{T}C_{M}|z$ that is associated with the current one. Hence, delta CoM potential energy can be defined as
\begin{equation*}
\Delta{E}_{P} = {E}_{P} - {E}_{P}^{0} = Mg\Delta C_{M}|z = Mg(C_{M}|z - C^{0}_{M}|z),
\end{equation*}
\noindent where $M$ is the mass of the subject, $g$ is the gravity acceleration, and $ \Delta C_{M}|z $ is the variation of the $z$-coordinate of the CoM, with $C^{0}_{M}|z$ the CoM in the neutral posture and $C_{M}|z$ the CoM in the current one.
Such a quantity provides the degree of deviation of the posture of the subjects from a neutral and convenient body configuration. Hence, to some extent, it is capable to monitor the risk associated with awkward and unfavourable postures.
Accordingly, the seventh index ${\bm\omega}_{7}$ is the normalised delta CoM potential energy, where $\Delta C_{M}|z$ is the displacement of the CoM height ($z$-coordinate) and $\Delta CoM_{z}^{max}$ is its maximum value, which can be estimated experimentally, by asking the subjects to move along the $z$-coordinate to the maximum extent possible and recording the maximum CoM displacements achieved. ${\bm\omega}_{7}$ is the only joint-independent index.

Finally, the eighth index ${\bm{\omega}}_{8}(\mathbf{q})$ derives from the ``compressive force'' concept that was originally conceived in \cite{fortini2020compressive} to account for the balanced inward (``pushing'') forces that result from the interaction of the human with the environment or objects. The aim of this study was to overcome the limitations of the overloading joint torque index. In fact, there are certain body configurations or task-dependent force profiles (magnitude and direction) that are characterised by negligible joint torques but considerable compressive forces, which may be harmful to the musculoskeletal system. 
Additionally, the compressive force index considers all the components of the external force. Details of the method can be found in \cite{fortini2020compressive}.
Based on the above, the eighth index ${\bm{\omega}}_{8}(\mathbf{q})$ is the joint normalised compressive force where $\mathbf{f}_{C}\in \mathbb{R}^{n_{j}}$ is the compressive force vector and $\mathbf{f}_{C}^{\text{max}}$ is the vector including the corresponding maximum values, which can be found experimentally, by asking the subjects to exert the maximum force possible and recording the maximum values achieved.

\section{Experimental analysis}
\label{sec:experiments}

To investigate the proposed monitoring framework as a tool to assess human ergonomics in the workplace, an experimental analysis was conducted on human subjects. Three different tasks were performed by each participant with the aim to simulate, in the laboratory settings, activities that are commonly carried out by workers in the current industrial scenario. Such tasks were selected to encompass the most significant risk factors in the workplace: mechanical overloading of the body joints, variable and high-intensity interaction forces, and repetitive and monotonous movements. Accordingly, lifting/lowering of a heavy object (task $1$), drilling (task $2$), and painting with a lightweight tool (task $3$), were considered, respectively, in this study.  
While the subjects were carrying out such activities, the data regarding the whole-body motion and the forces exchanged with the environment (both GRF and interaction forces at the end-effector) were collected and the full set of ergonomic indexes (see Table \ref{tab:indexes}) was estimated. In addition, the ergonomic risk scores resulting from the EAWS \cite{schaub2013european} was computed in an off-line phase for the sake of comparison. The corresponding analysis and computations are carried out in collaboration with Fondazione Ergo (headquarters: Varese, Italia). 
The whole experimental procedure was carried out at Human-Robot Interfaces and Physical Interaction (HRII) Lab, Istituto Italiano di Tecnologia (IIT) in accordance with the Declaration of Helsinki, and the protocol was approved by the ethics committee Azienda Sanitaria Locale (ASL) Genovese N.3 (Protocol IIT\_HRII\_ERGOLEAN 156/2020).

\subsection{Experimental setup}

\begin{figure*}[h]
	\centering
	\begin{subfigure}[c]{0.37\textwidth}
		\centering
		\includegraphics[width=0.9\textwidth]{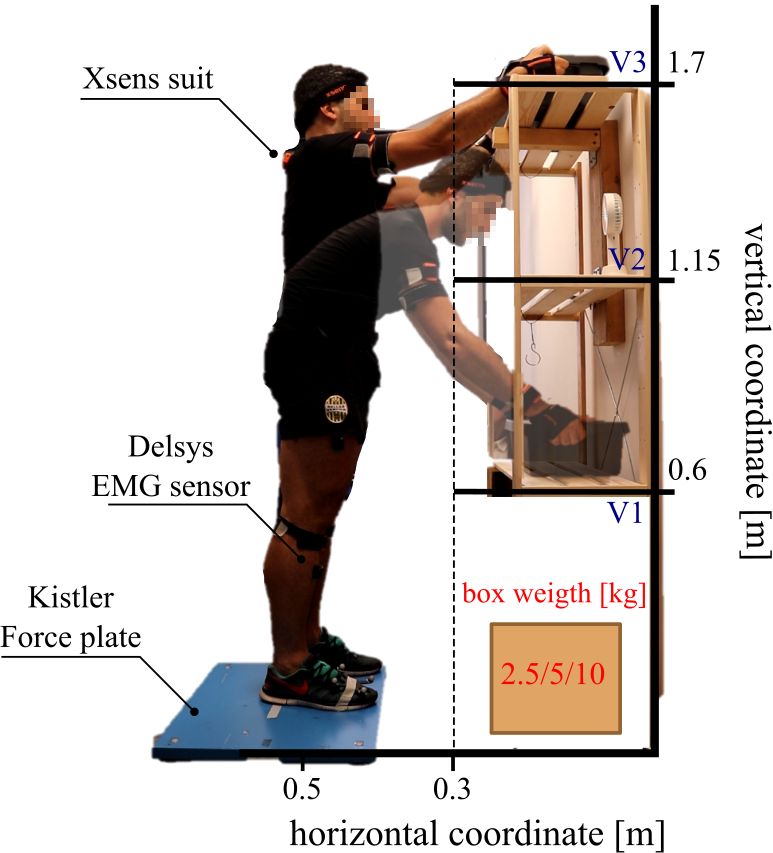}
		\caption{Task $1$}
		\label{fig:scenario001}
	\end{subfigure}
	\begin{subfigure}[c]{0.28\textwidth}
		\centering
		\includegraphics[trim = 0 10 0 30 mm,clip,width=0.9\textwidth]{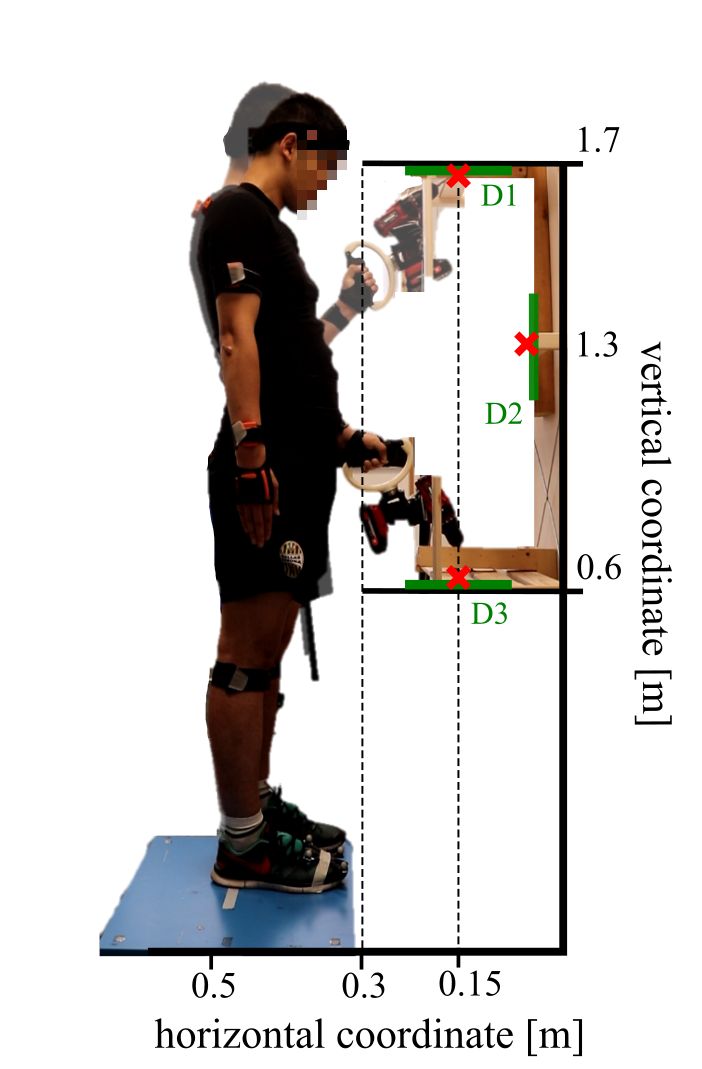}
		\caption{Task $2$}
		\label{fig:scenario002}
	\end{subfigure}
	\begin{subfigure}[c]{0.31\textwidth}
		\centering
		\includegraphics[trim = 0 20 0 0 mm,clip,width=0.9\textwidth]{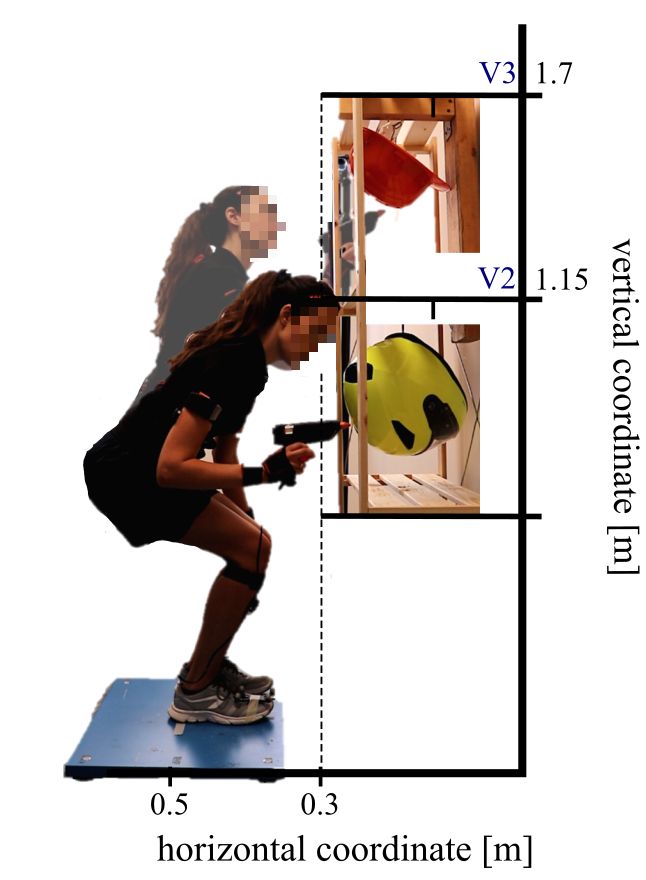}
		\caption{Task $3$}
		\label{fig:scenario003}
	\end{subfigure}
	\caption{Experimental setups to investigate the proposed monitoring framework as a tool to assess human ergonomics while performing three typical working activities in the current industrial scenario: \textbf{(a)} lifting/lowering of a heavy object (task $1$), \textbf{(b)} drilling (task $2$), \textbf{(c)} and painting with a spray gun (task $3$). The sensor systems employed are also highlighted.}
	\label{fig:scenarios}
	\vspace{-5 mm} 
\end{figure*}

Twelve healthy volunteers, eight males and four females, (age: $28.9 \pm 3.5$ years; mass: $69.3 \pm 13.5$ kg; height: $173.9 \pm 6.8$ cm)\footnote{Subject data is reported as: mean $\pm$ standard deviation.} were recruited in the experimental analysis. Participants were students and researchers with no or limited experience of industrial work. Written informed consent was obtained after explaining the experimental procedure and a numerical ID was assigned to anonymise the data. 

The experimental setups for the three selected tasks are illustrated in Figure \ref{fig:scenarios}. The sensor systems employed are also highlighted.
The subjects were required to wear the Xsens MVN Biomech suit, an inertial-based motion-capture suit, commercialised by Xsens Technologies B.V. (headquarters: Enschede, Netherlands) to track the whole-body motion and to stand on the Kistler force plate, commercialised by Kistler Holding AG (headquarters: Winterthur, Switzerland) to measure the whole-body CoP and GRF.
In addition, a sEMG system, the Delsys Trigno Wireless platform, commercialised by Delsys Inc. (headquarters: Natick, MA, United States), was employed to measure muscle activity as a reference to the effective physical effort required for the tasks. Specifically, ten muscles were analysed: the Anterior Deltoid (AD), the Posterior Deltoid (PD), the Biceps Brachii (BC), the Triceps Brachii (TC), the Trapezius Descendens (TR), the Erector Spinae (ES), the Gluteus Maximus (GM), the Rectus Femoris (RF), the Biceps Femoris (BF) and the Tibialis Anterior (TA). The subjects were required to bound their movements to the sagittal plane since a human sagittal model was adopted in this study. Accordingly, the selection of the muscles was made considering the ones more involved in flexion/extension movements. Prior to the experiment, the maximum voluntary contraction (MVC) exerted by each muscle was recorded for all the participants with the aim to normalise the muscle activity. The acquisition and synchronization of all the sensor data are managed using Robot Operating System (ROS) environment.

All the subjects were required to perform all three tasks. No specific instructions were provided to the subjects about the way to perform the task and no preliminary learning phase was conducted. For each task, they were asked to perform three trials and between one trial and the following, there was a resting period of $2$ minutes. 

Task $1$ consisted in lifting/lowering a box on a shelf and put it/take it at/from three different height levels w.r.t. floor that were located under the human knees (V1), at the level of the pelvis (V2) and over the shoulder (V3), respectively (see Figure \ref{fig:scenario001}). Indeed, such an activity induces high-intensity forces and torques within the human body that may lead to the mechanical overload of biological structures such as muscles and joints.
Since the detrimental effect of mechanical overload primarily depends on the magnitude of the force required to accomplish it, three boxes of different weights (thus demanding different effort) were considered. Specifically, the weights of the boxes were $2.5$, $5.0$, and $10.0$ kg, respectively. During a trial, all the possible combinations of box weight/height level were considered in a randomized order to counter-balance the effect of fatigue on the participants' muscles. 

Even though the load affecting the musculoskeletal system during a job depends mainly on the magnitude of the force exerted/experienced by workers, other factors can contribute to its intensification. Namely: the direction of the force, its variability over time, and the postural demands required to develop it. Accordingly, task $2$ consisted in drilling three wood panels that are located in three different positions (D1, D2, D3) on a shelf, highlighted in green in Figure \ref{fig:scenario002}. The holes, made with the driller, had to be performed approximately in the points where the red crosses are depicted.
During a trial, each point had to be drilled for $10.0$ s, in the sequence \footnote{It should be noted that the randomized order was considered only for task $1$ that was longer and more demanding. For tasks $2$ and $3$, the order was the same for all the trials since the effect of fatigue was assumed negligible.} D1, D2, D3, with a break of approximately $5.0$ s in between. Throughout the whole duration of the task, the interaction forces were measured via the developed hand/tool interface. For this purpose, the subjects had to keep the driller through a dedicated handle, which was connected to a Mini45 F/T sensor, commercialised by ATI Industrial Automation (headquarters: Apex, NC, United States). 
The weight of the tool (driller and handle) was $2.5$ kg.

As mentioned above, repetitive actions with lightweight objects may result in the accumulation of local muscle fatigue, which
can cause severe injuries \cite{shin2007measurement}.
Hence, the last task selected was spray painting with a lightweight tool. Task $3$ consisted in painting two helmets along their whole surface 
with a spray gun. The helmets were hanging from a hook at two different height levels (V2 and V3, respectively). During a trial, the subjects had to spray each helmet (from V3 to V2 position) moving smoothly and slowly the spray gun in proximity to the helmet: from top to bottom for $10$ s, then from to bottom to top for $10$ s and finally again from top to bottom for $10$ s. In total, each helmet has to be painted for $30$ s, with a break of $5$ s approximately in between.

After the full set of ergonomic indexes was estimated, a statistical method, i.e. the analysis of variance (ANOVA), was employed to verify the significance of the results, considering a level of statistical significance equal to $0.05$. Separate one-way repeated measures ANOVAs were applied to determine the capability of each ergonomic index to discriminate between different load levels (i.e. experimental conditions). Pairwise comparisons were also conducted by using post-hoc paired t-tests. The analysis was performed separately for each considered joint. 
Finally, the sEMG signals, which were collected as a benchmark for the human physical effort, were first filtered (passband: 2-500 Hz) to remove movement artifacts for all the ten muscles \cite{stegeman2007standards}. Then, they were rectified and normalised using the MVC values previously recorded for each subject.

\subsection{Results}
\label{sec:results}

In this section, the results of the simulation experiments of typical working activities are presented for all the three selected tasks performed by twelve subjects. In addition, the findings of the analysis carried out using the EAWS method are provided.

\subsubsection{Working activities simulation experiments}

\begin{figure}[!h]
	\centering
	\begin{subfigure}[c]{0.13\textwidth}
		\centering
		\caption{2.5 kg, V1}
		\includegraphics[trim= 0 0 0 0 mm, clip,width=\textwidth]{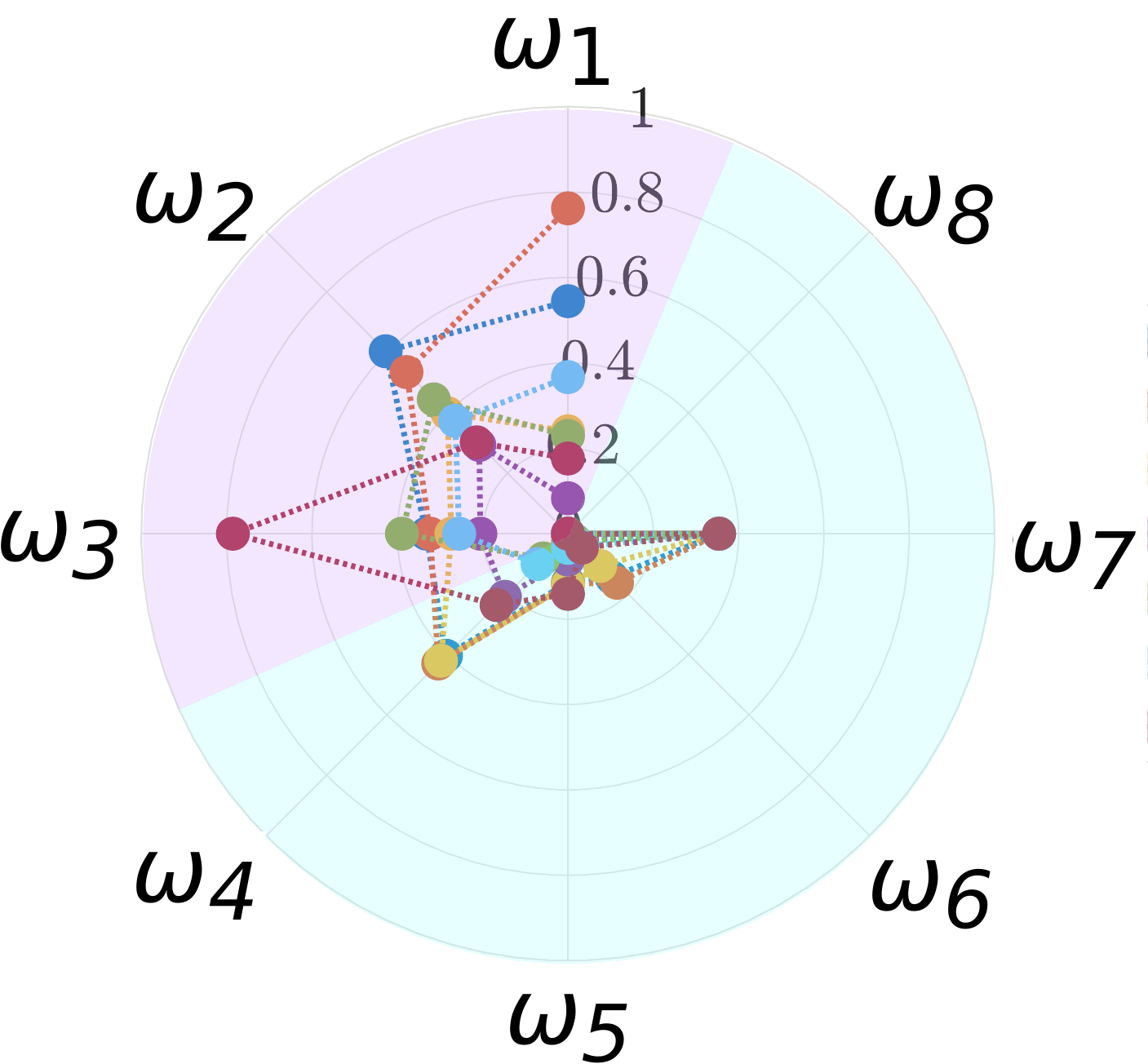}
	\end{subfigure}\hspace{5 mm}
	\begin{subfigure}[c]{0.13\textwidth}
		\centering
		\caption{2.5 kg, V2}
		\includegraphics[trim= 0 0 0 0 mm, clip,width=\textwidth]{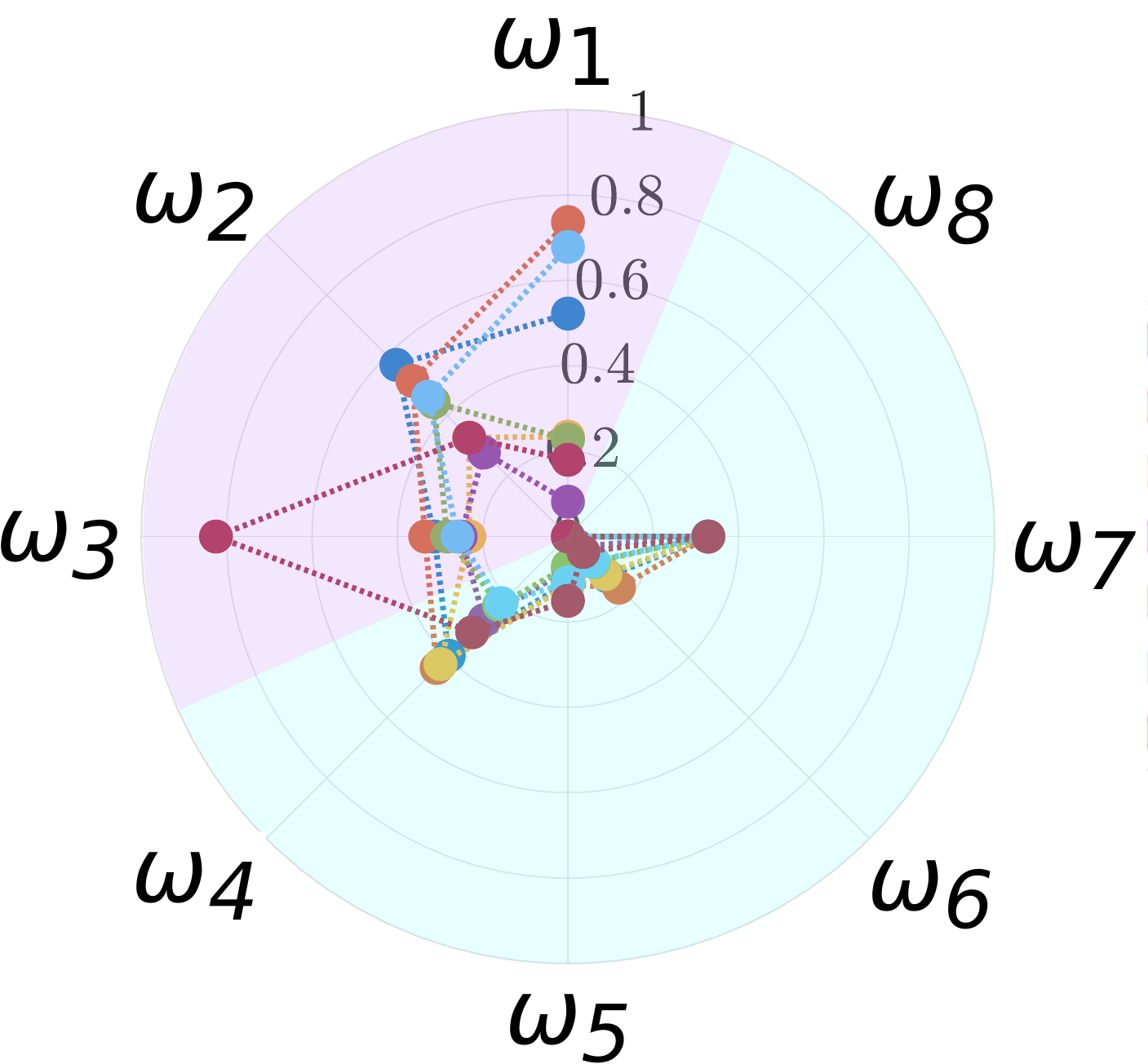}
	\end{subfigure}\hspace{5 mm}
	\begin{subfigure}[c]{0.13\textwidth}
		\centering
		\caption{2.5 kg, V3}
		\includegraphics[trim= 0 0 0 0 mm, clip,width=\textwidth]{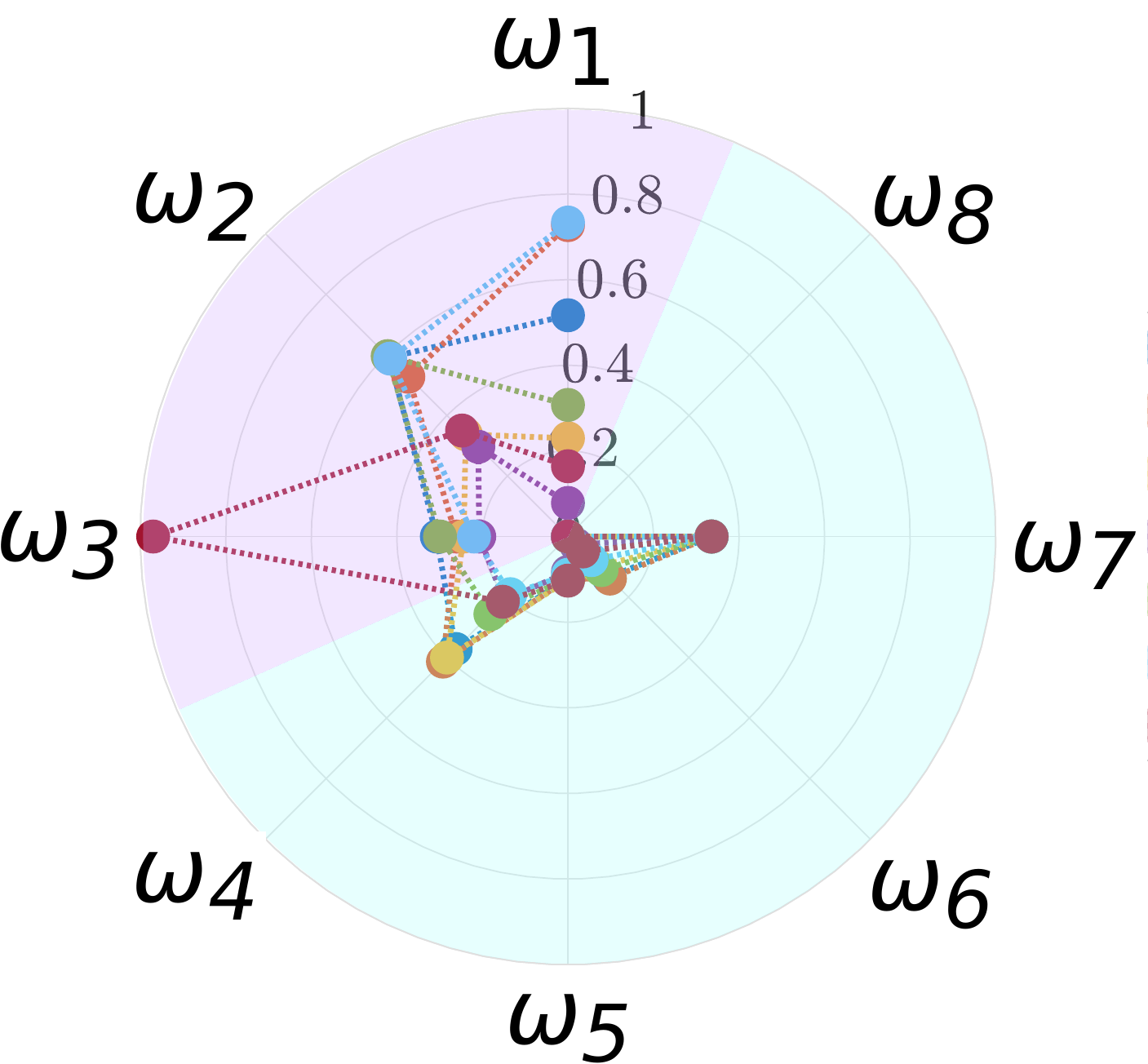}
	\end{subfigure}
	\begin{subfigure}[c]{0.13\textwidth}
		\centering
		\vspace{5 mm}
		\caption{5.0 kg, V1}
		\includegraphics[trim= 0 0 0 0 mm, clip,width=\textwidth]{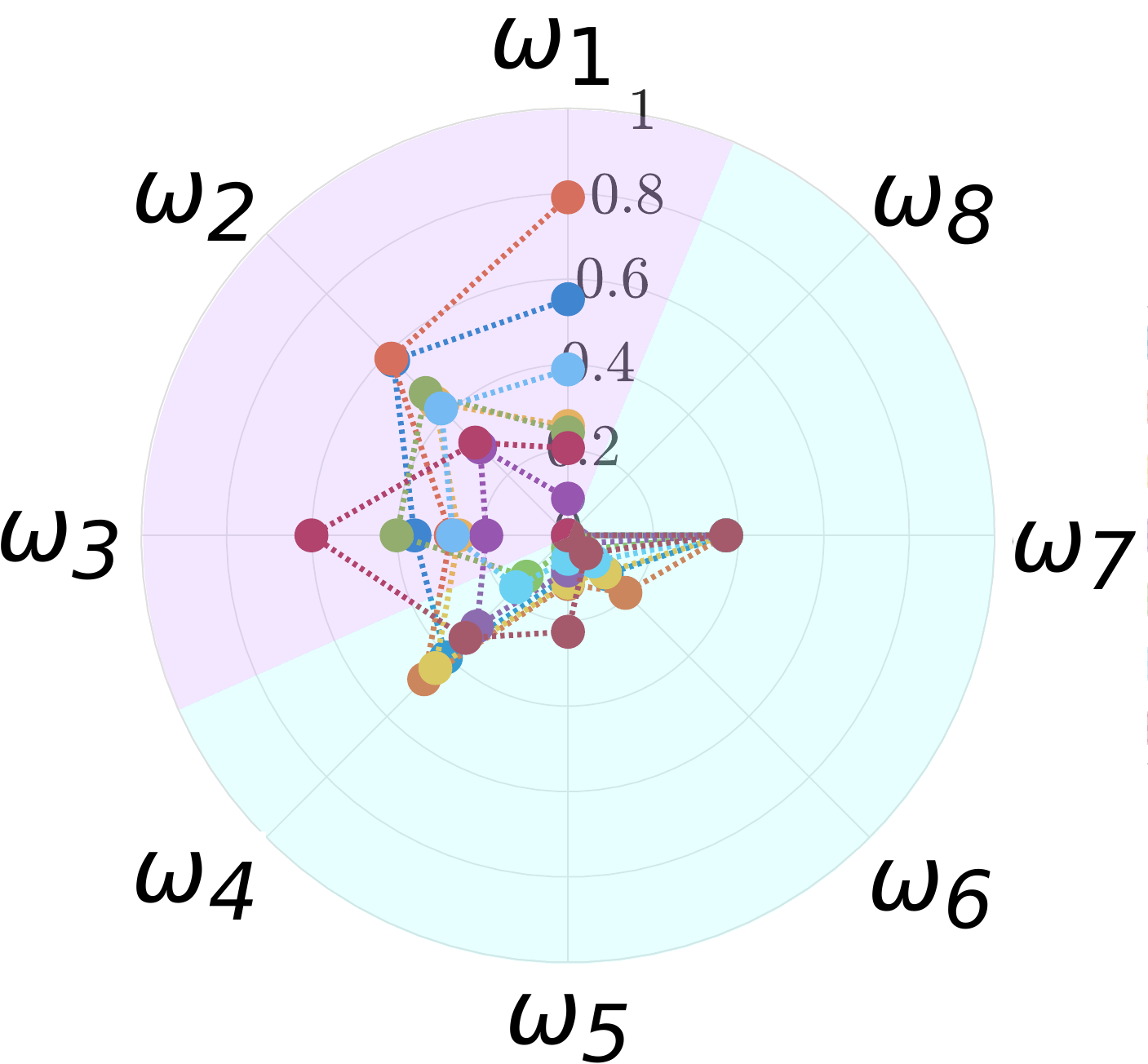}
	\end{subfigure}\hspace{5 mm}
	\begin{subfigure}[c]{0.13\textwidth}
		\centering
		\vspace{5 mm}
		\caption{5.0 kg, V2}
		\includegraphics[trim= 0 0 0 0 mm, clip,width=\textwidth]{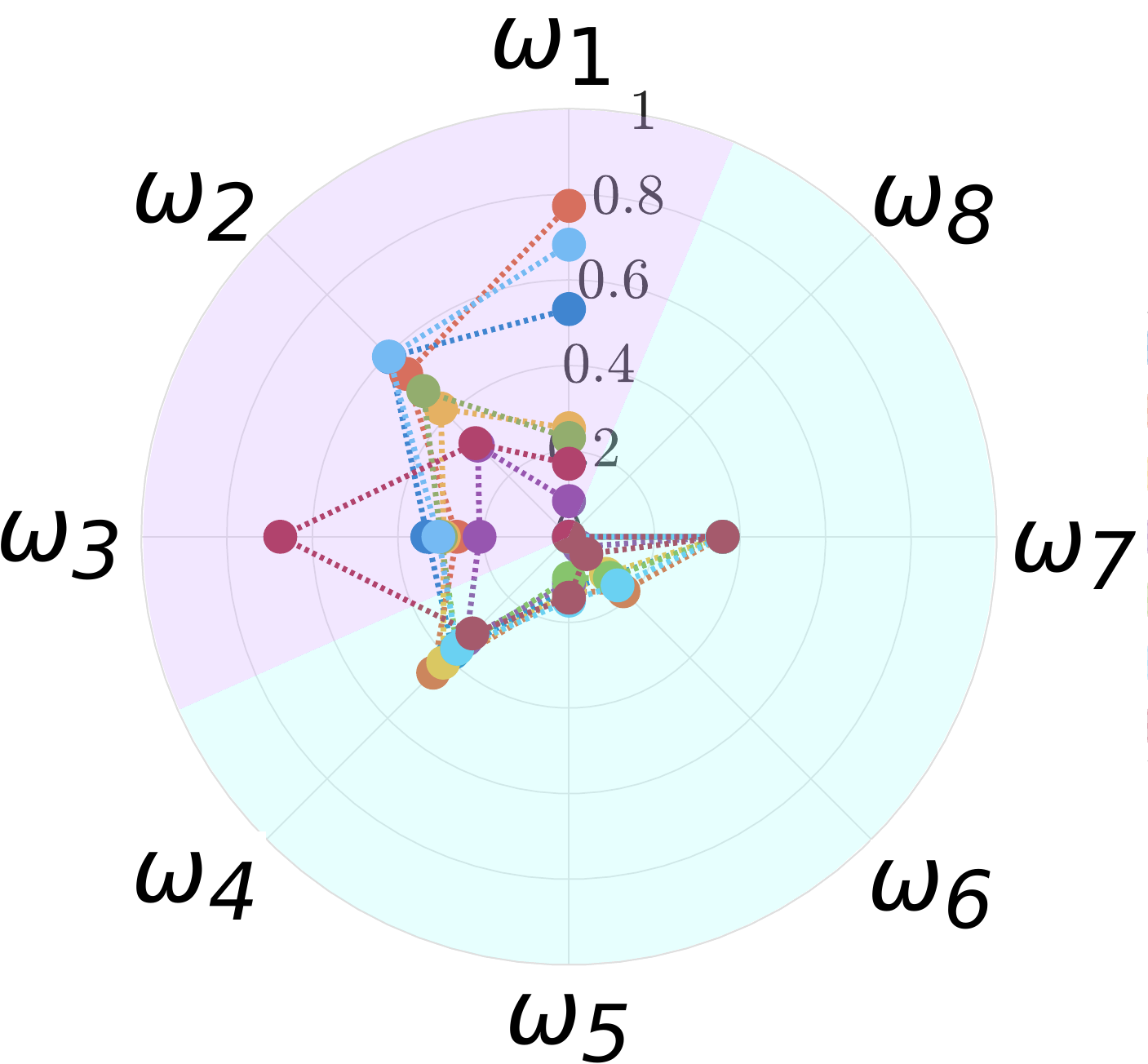}
	\end{subfigure}\hspace{5 mm}
	\begin{subfigure}[c]{0.13\textwidth}
		\centering
		\vspace{5 mm}
		\caption{5.0 kg, V3}
		\includegraphics[trim= 0 0 0 0 mm, clip,width=\textwidth]{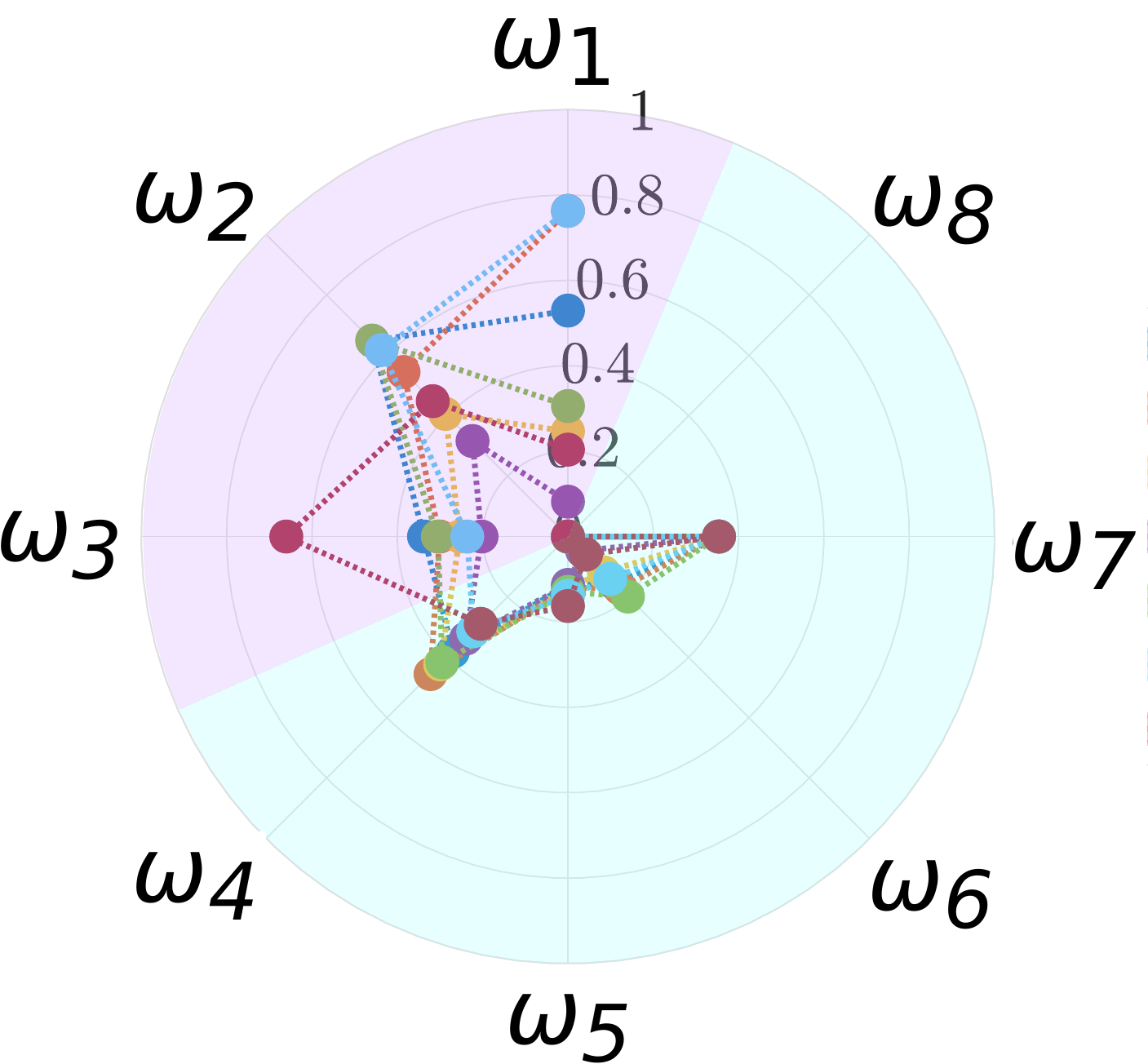}
	\end{subfigure}
	\begin{subfigure}[c]{0.13\textwidth}
		\centering
		\vspace{5 mm}
		\caption{10.0 kg, V1}
		\includegraphics[trim= 0 0 0 0 mm, clip,width=\textwidth]{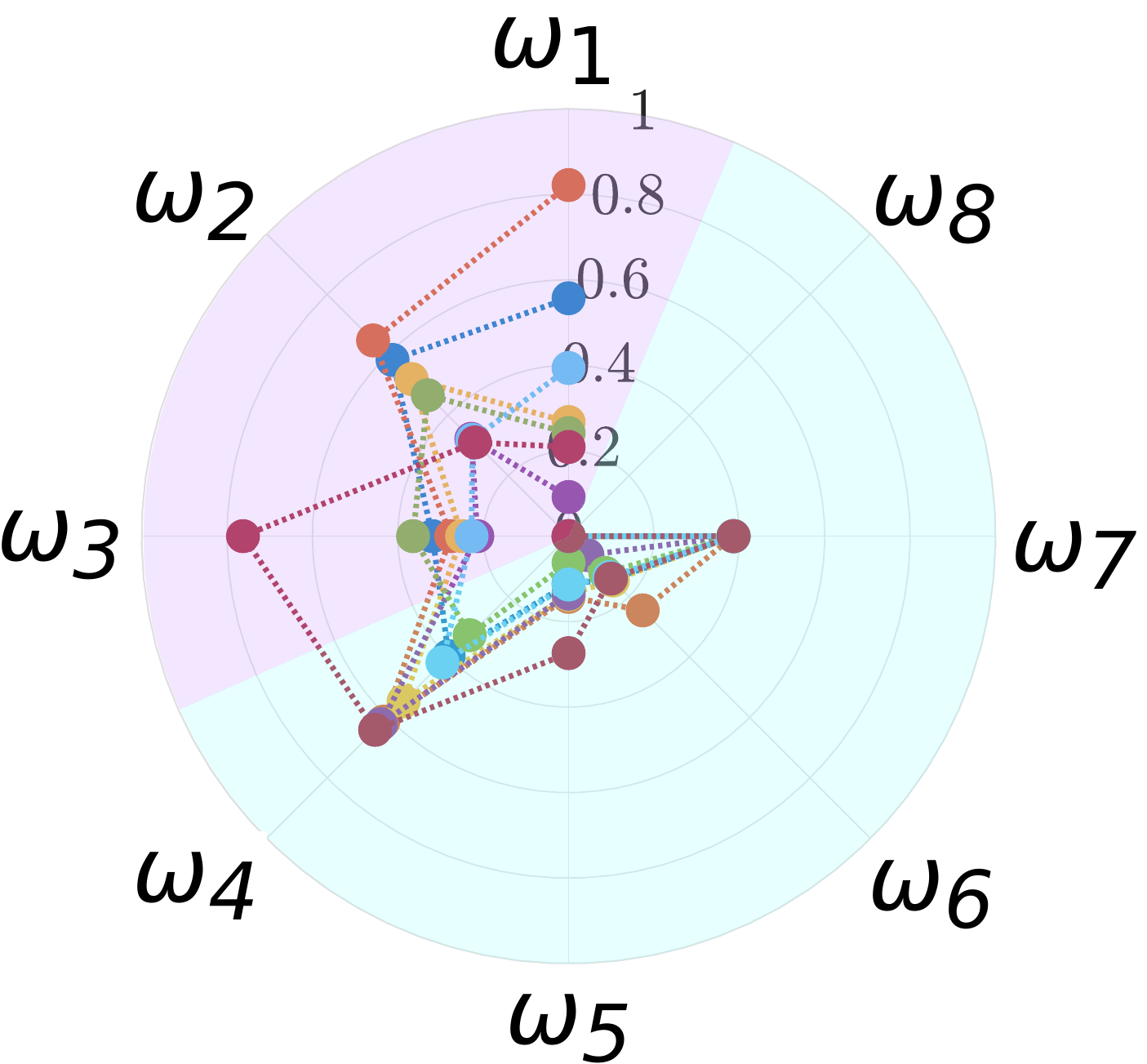}
	\end{subfigure}\hspace{5 mm}
	\begin{subfigure}[c]{0.13\textwidth}
		\centering
		\vspace{5 mm}
		\caption{10.0 kg, V2}
		\includegraphics[trim= 0 0 0 0 mm, clip,width=\textwidth]{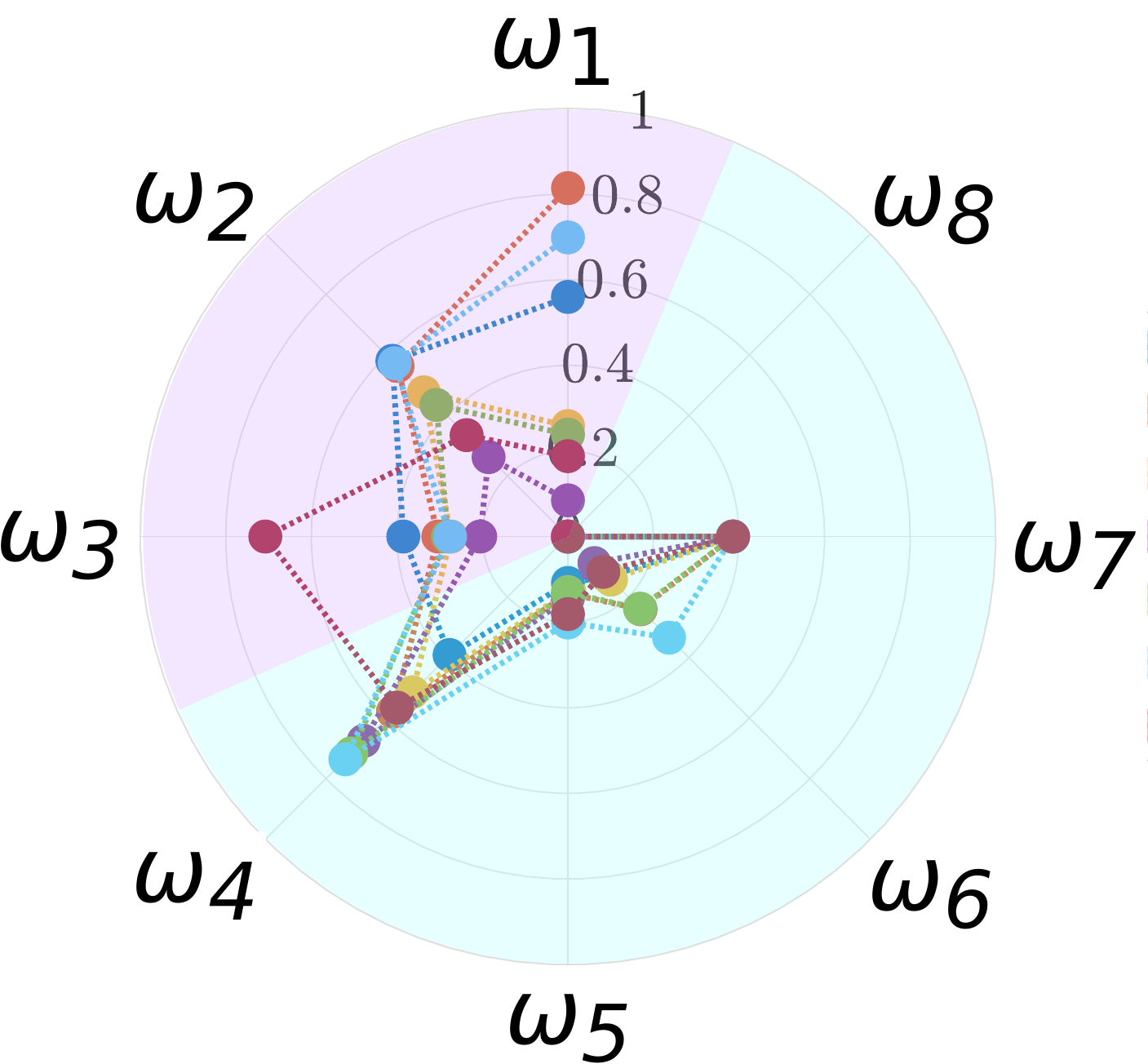}
	\end{subfigure}\hspace{5 mm}
	\begin{subfigure}[c]{0.13\textwidth}
		\centering
		\vspace{5 mm}
		\caption{10.0 kg, V3}
		\includegraphics[trim= 0 0 0 0 mm, clip,width=\textwidth]{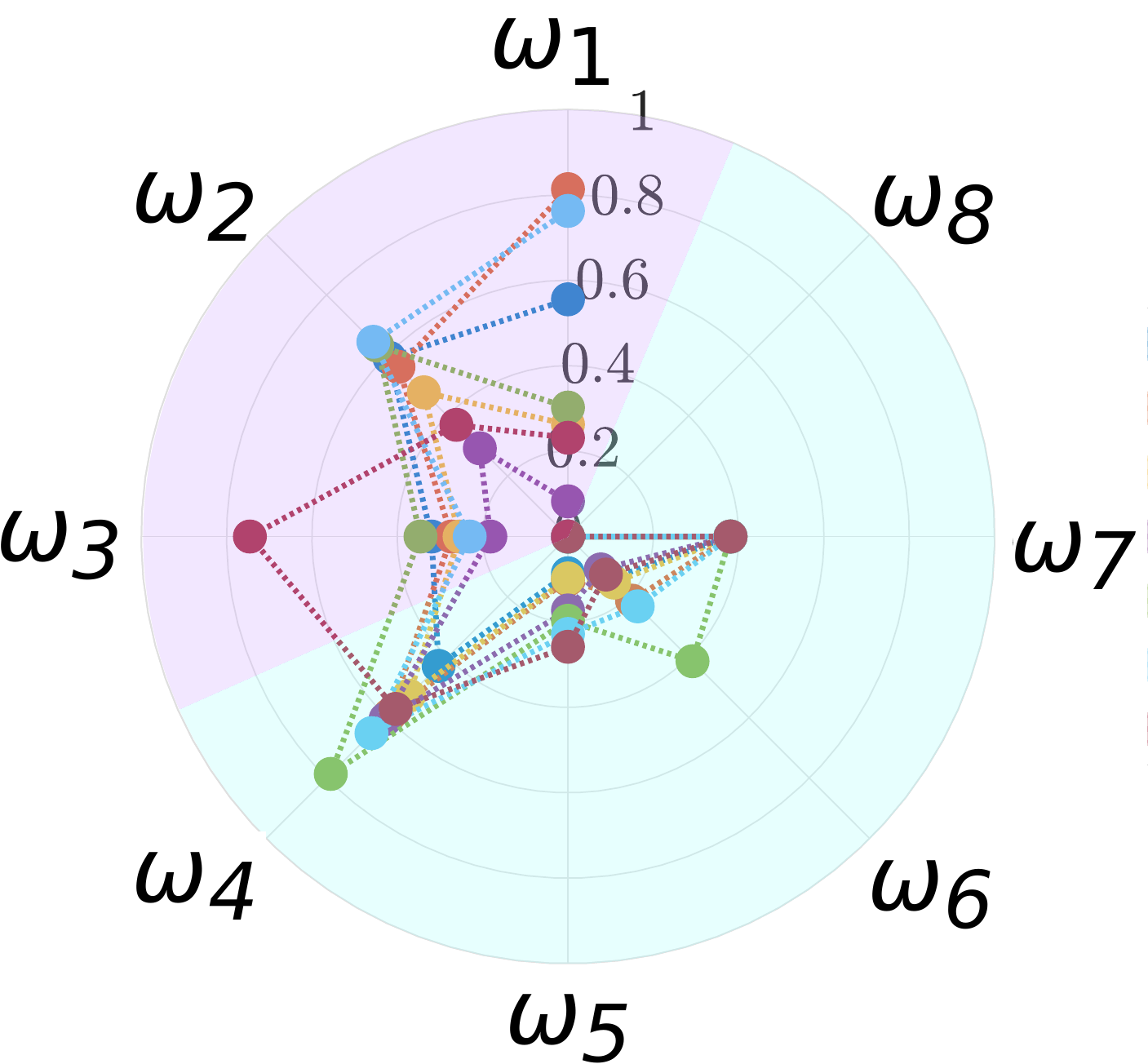}
	\end{subfigure}
	\begin{subfigure}[c]{0.4\textwidth}
		\centering
		\vspace{5 mm}
		\includegraphics[trim= 0 0 0 0 mm, clip,width=\textwidth]{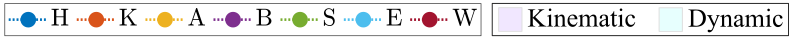}
	\end{subfigure}
	\vspace{2 mm}
	\caption{Polar plots to represent the overall outcome of the proposed ergonomic indexes for task $1$. For each experimental condition, i.e. box weight ($2.5$, $5.0$, $10.0$ kg) and height level (V1, V2, V3), the maximum values of each index over the trial are shown in the human main joints: hip (H), knee (K), ankle (A), back (B), shoulder (S), elbow (E) and wrist (W).}
	\label{fig:Task001_polar}
	\vspace{-8mm}
\end{figure}

\begin{figure}[!h]
	\centering
	\begin{subfigure}[c]{0.13\textwidth}
		\centering
		\caption{D1}
		\includegraphics[trim= 0 0 0 0 mm, clip,width=\textwidth]{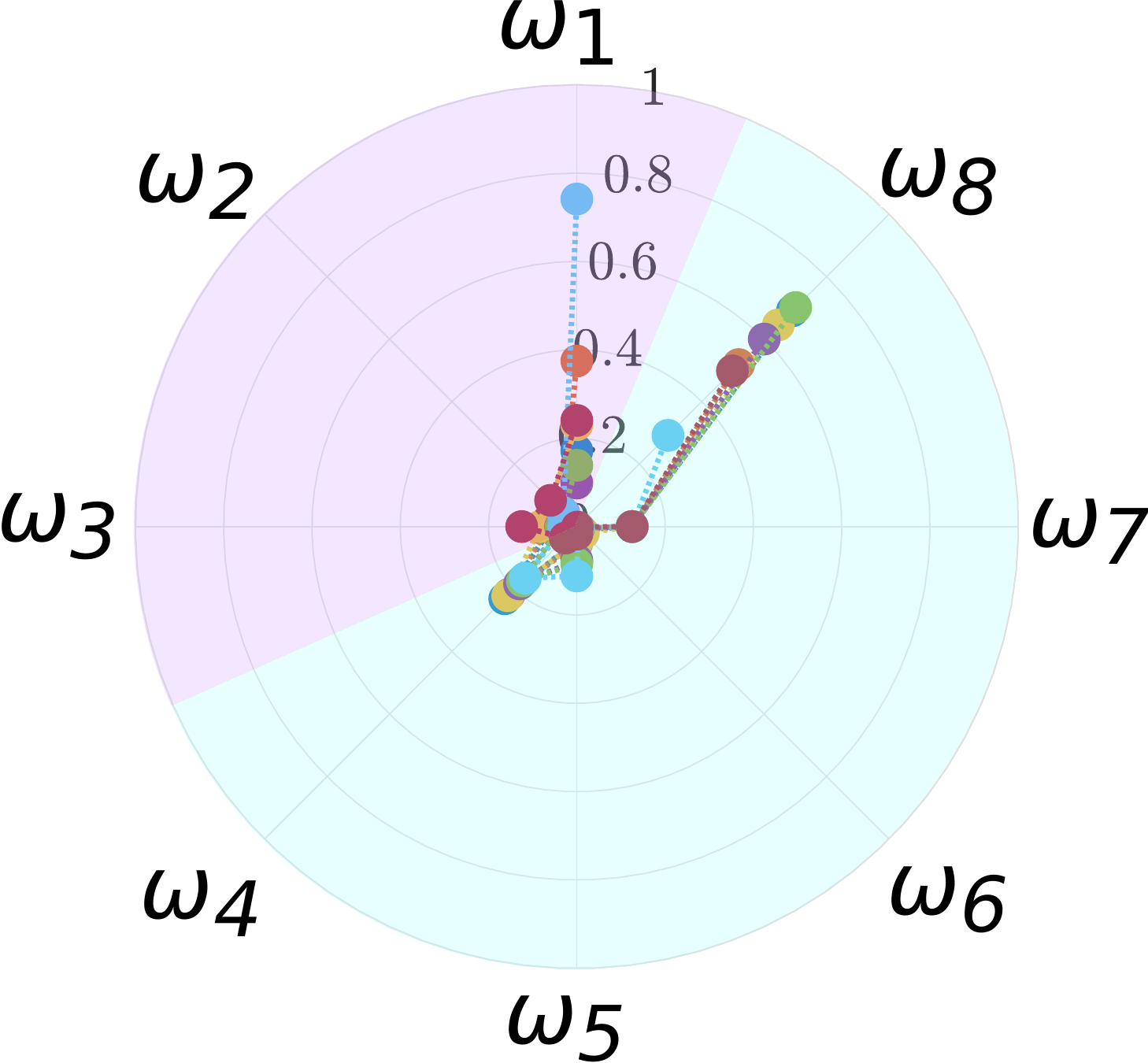}
	\end{subfigure}\hspace{5 mm}
	\begin{subfigure}[c]{0.13\textwidth}
		\centering
		\caption{D2}
		\includegraphics[trim= 0 0 0 0 mm, clip,width=\textwidth]{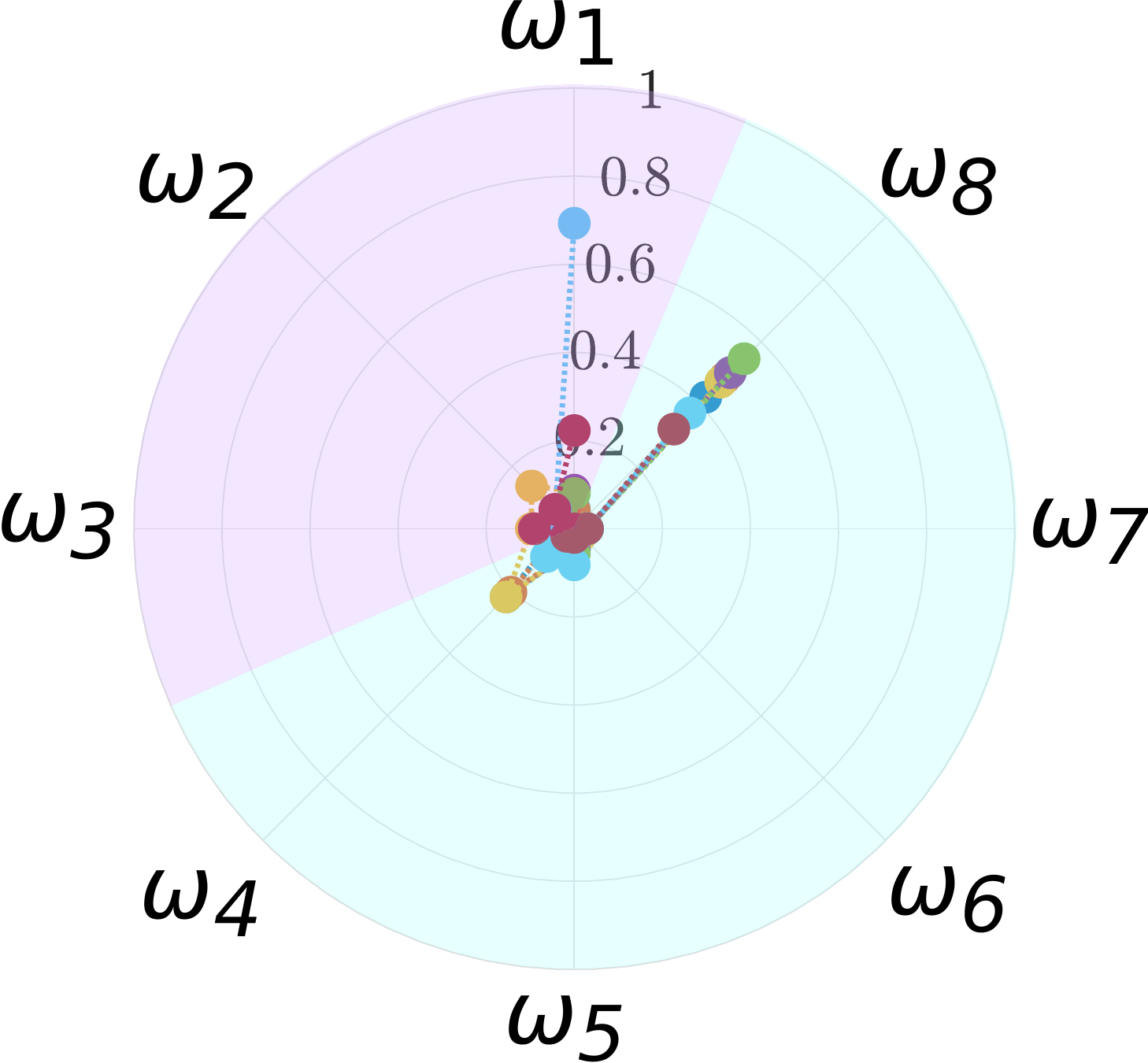}
	\end{subfigure}\hspace{5 mm}
	\begin{subfigure}[c]{0.13\textwidth}
		\centering
		\caption{D3}
		\includegraphics[trim= 0 0 0 0 mm, clip,width=\textwidth]{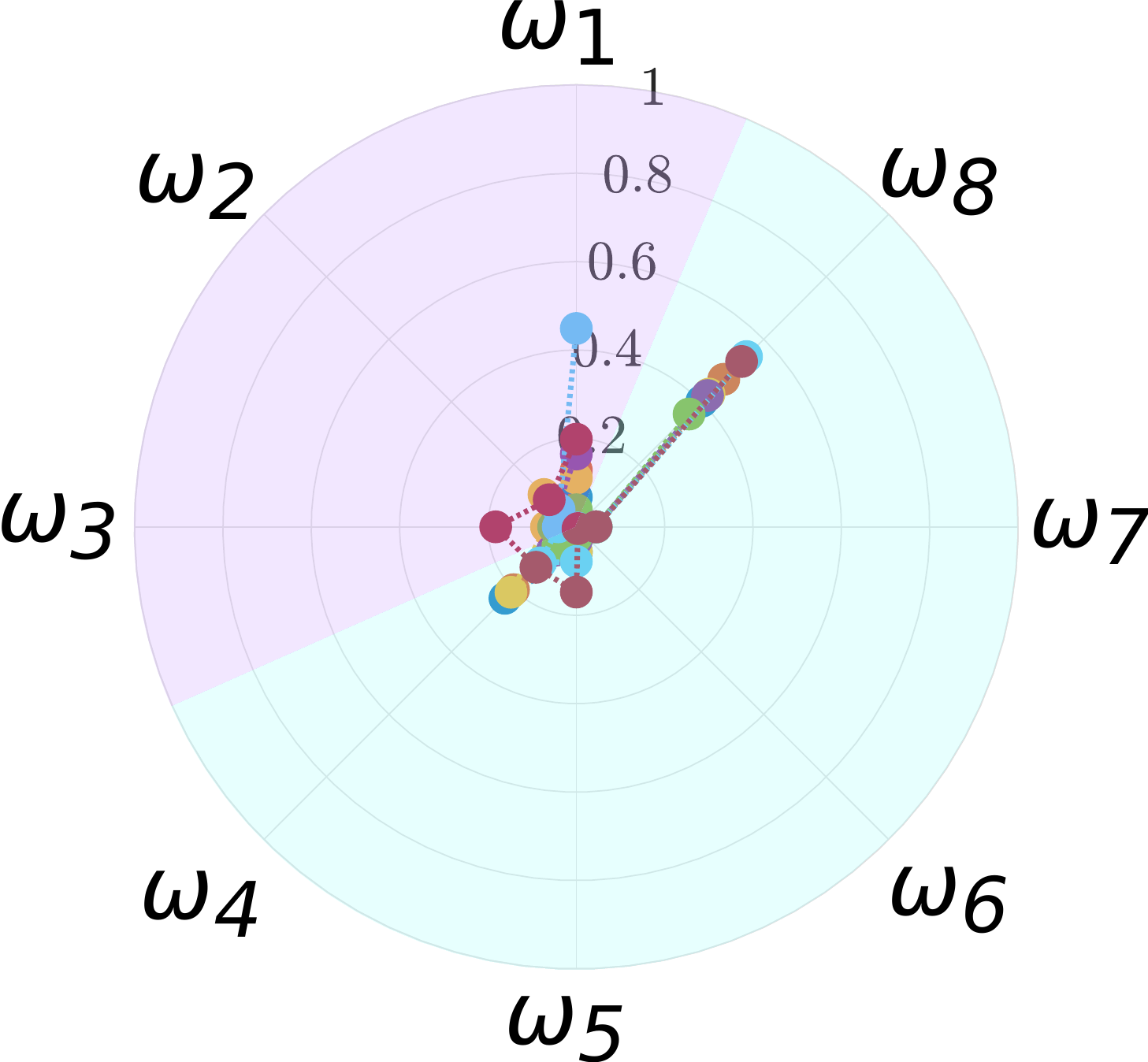}
	\end{subfigure}
	\begin{subfigure}[c]{0.4\textwidth}
		\centering
		\vspace{5 mm}
		\includegraphics[trim= 0 0 0 0 mm, clip,width=\textwidth]{Figure14.png}
	\end{subfigure}
	\vspace{2 mm}
	\caption{Polar plots to represent the overall outcome of the proposed ergonomic indexes for task $2$. For each experimental condition, i.e. the positions (D1, D2, D3) of the panels to be drilled, the maximum values of each index over the trial are illustrated in the human main joints: hip (H), knee (K), ankle (A), back (B), shoulder (S), elbow (E) and wrist (W).}
	\label{fig:Task002_polar}
\end{figure}

\begin{figure}[!h]
	\centering
	\begin{subfigure}[c]{0.22\textwidth}
		\centering
		\caption{First helmet}
		\includegraphics[trim= 0 0 0 0 mm, clip,width=\textwidth]{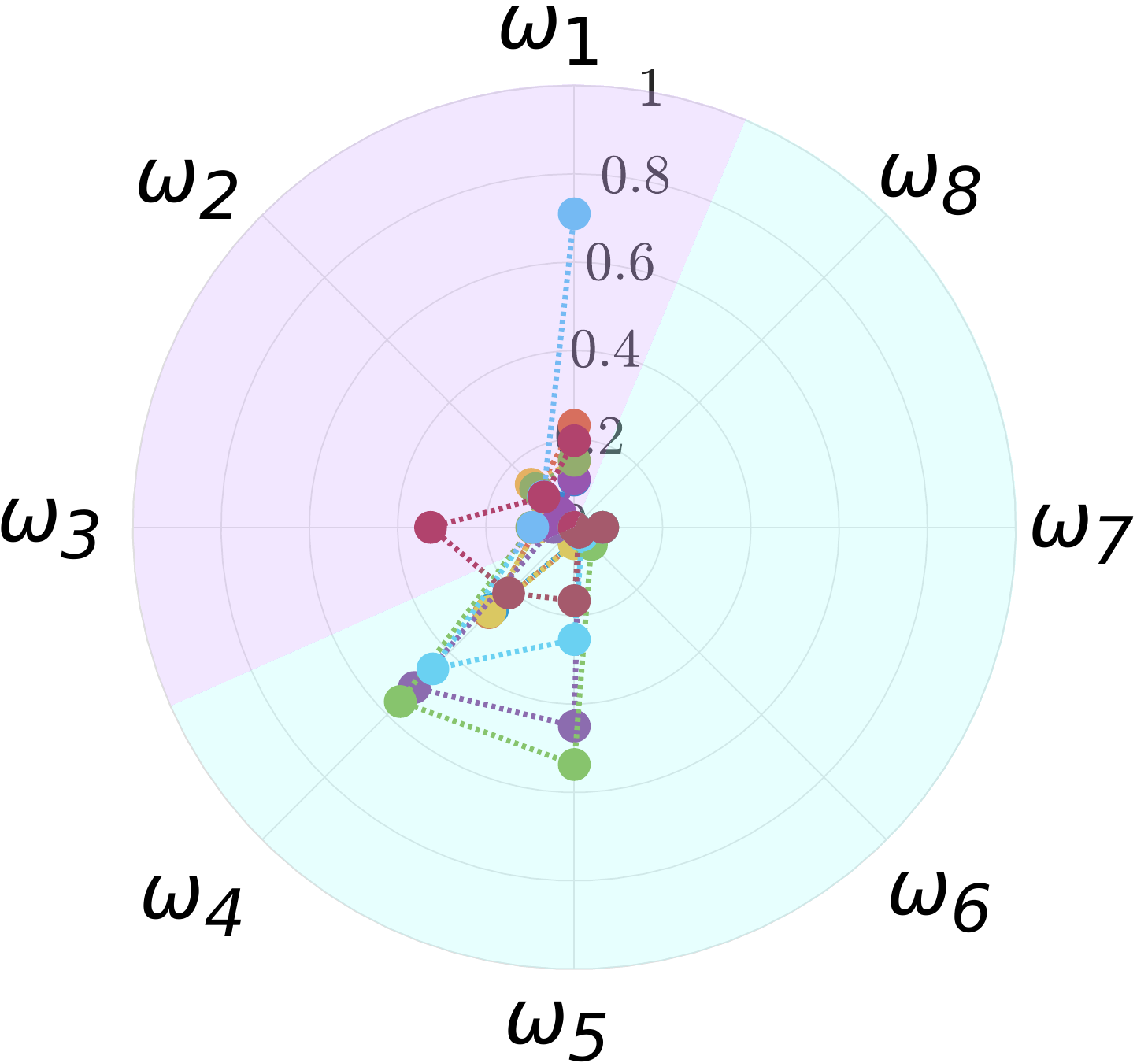}
	\end{subfigure}\hspace{5 mm}
	\begin{subfigure}[c]{0.22\textwidth}
		\centering
		\caption{Second helmet}
		\includegraphics[trim= 0 0 0 0 mm, clip,width=\textwidth]{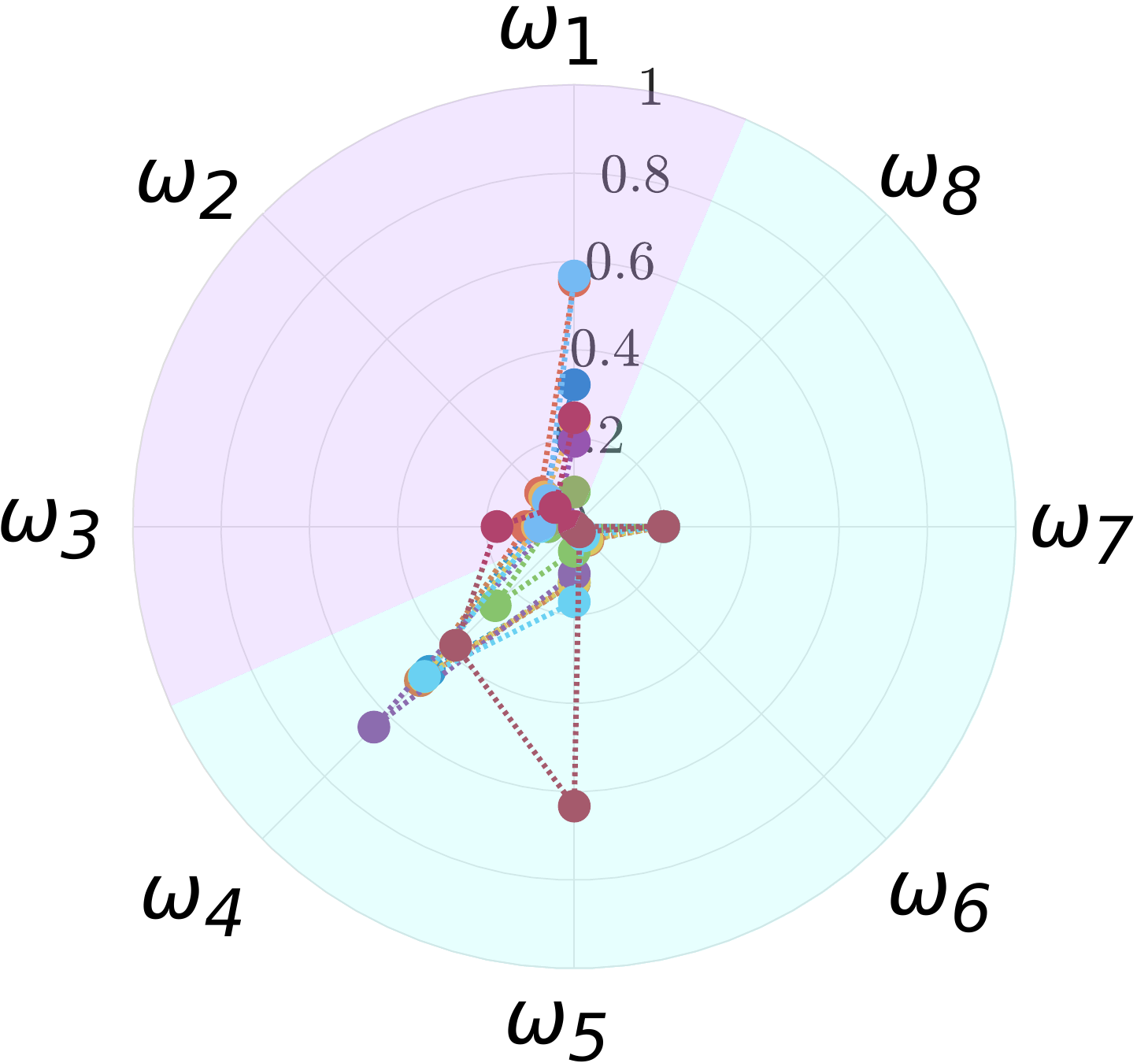}
	\end{subfigure}
	\begin{subfigure}[c]{0.4\textwidth}
		\centering
		\vspace{5 mm}
		\includegraphics[trim= 0 0 0 0 mm, clip,width=\textwidth]{Figure14.png}
	\end{subfigure}
	\vspace{2 mm}
	\caption{Polar plots to represent the overall outcome of the proposed ergonomic indexes for task $3$. For each phase of the task, i.e. one of the two helmets to be painted, the maximum values of each index over the phase are illustrated in the human main joints: hip (H), knee (K), ankle (A), back (B), shoulder (S), elbow (E) and wrist (W).}
	\label{fig:Task003_polar}
    \vspace{-2 mm}
\end{figure}

The outcome of the proposed ergonomic indexes has been resumed in the aggregated polar plots illustrated in Figures \ref{fig:Task001_polar}, \ref{fig:Task002_polar} and \ref{fig:Task003_polar} for task $1$, task $2$ and task $3$, respectively. For task $1$, a plot is provided for each one of the nine actions corresponding to a specific experimental condition (box weight, height level). For task $2$, a plot is provided for each one of the three actions corresponding to the three different positions (D1, D2, D3) of the panels to be drilled. For task $3$, two plots are considered to represent the two phases, i.e. the two helmets to be processed, of the painting activity. 
Each plot includes eight axes to correspond to the eight ergonomic indexes listed in Table \ref{tab:indexes}. For each index, the maximum value of each action/phase is exhibited for all the considered joints, specifically: hip (H), knee (K), ankle (A), back (B), shoulder (S), elbow (E) and wrist (W). It should be noted that the choice of the maximum value is due to the short duration of the task, however, also other values such as the root mean square could have been considered. To highlight the scope of the ergonomic indexes, two separate areas are enhanced on the polar plots: the pink area encompasses the indexes accounting for kinematic aspects while the light blue area encompasses the indexes accounting for dynamic aspects.

\begin{figure}[h]
    \centering
    \vspace{-3 mm}
    \includegraphics[width=0.5\textwidth]{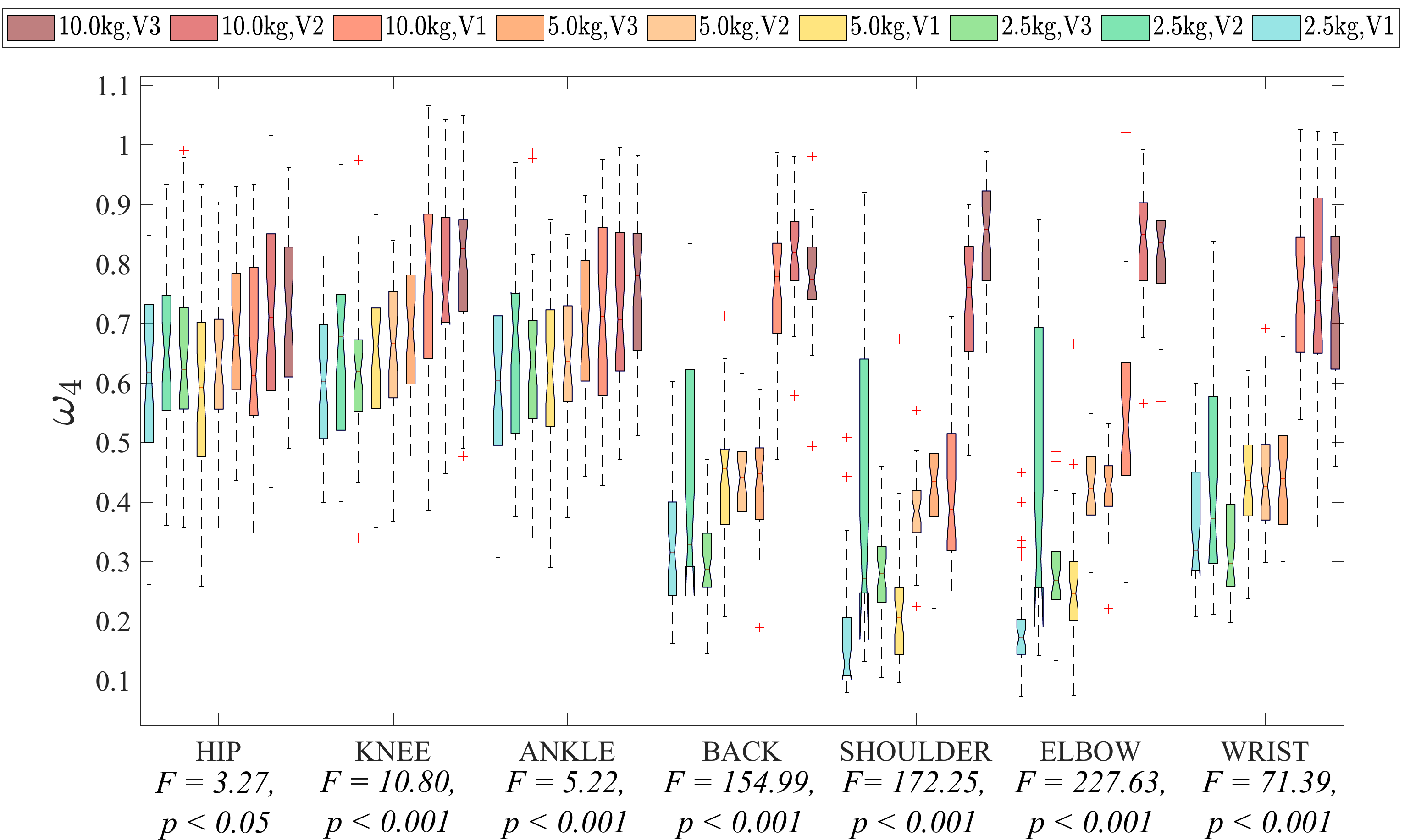}
    \caption{Boxplots for the overloading joint torque index $\bm\omega_{4}$ averaged among all the trials and subjects for the experimental conditions of task $1$, i.e. box weight ($2.5$, $5.0$, $10.0$ kg) and height level (V1, V2, V3), in the human main joints. The $F$-ratio ($F$) and $p$-values ($p$) resulting from the ANOVA are also reported.}
    \label{fig:boxplot_1}
\end{figure}

Let us consider the plots in Figure \ref{fig:Task001_polar}, relative to task $1$. Focusing on the pink area, which comprises the indexes accounting for the kinematic aspects, it can be noted that the highest values were reported by $\bm\omega_{1}$, namely, the joint displacement index. At the hip, knee, and elbow level the joint angles approached the boundaries of the RoM (i.e. mechanical joint limits) to a slightly different extent among the experimental conditions, suggesting that the subjects were required to adopt potentially unfavourable postures to execute this specific task. As regards $\bm\omega_{2}$, the values are gathered approximately in the middle of the range and no meaningful differences in joint velocities can be found among the different experimental conditions. Similarly, as shown by $\bm\omega_{3}$, comparable joint accelerations were exhibited by the subjects, even varying the conditions, with values in the lower half of the index range for all the joints. Interestingly, the only exception is the acceleration at the wrist level that presented rather high values. Since the variations of wrist acceleration have proven to correlate with ergonomic risk factors in the workplace \cite{estill2000use}, this data acquire certain importance for the purposes of the current analysis. On the other hand, focusing on the light blue area of the polar plots, which covers the indexes accounting for the dynamic aspects, it can be noted that the highest values were reported by $\bm\omega_{4}$, the overloading joint torque index. $\bm\omega_{5}$ and $\bm\omega_{6}$ remained generally low and steady among the different experimental conditions, $\bm\omega_{7}$ (which is actually joint-independent) was slightly higher but likewise without considerable variations. $\bm\omega_{8}$ could not just be estimated due to the absence of full knowledge about interaction forces for this task. Conversely, the overloading joint torque index exhibited a clear growth with increasing box weight and to a small extent with increasing height level.

A statistical analysis was also performed to compare the variance between experimental conditions on the dynamic ergonomic indexes. 
In Figure \ref{fig:boxplot_1}, the boxplots representing the values of $\bm\omega_{4}$ averaged among all the trials and subjects for all the experimental conditions of task $1$,
are illustrated in the human main joints. The $F$-ratio ($F$) and $p$-values ($p$) resulting from the ANOVA are also reported. For all the joints, significant differences can be observed among the different experimental conditions ($p$ $<0.05$ for the hip, $p$ $<0.001$ for the other joints). On the other hand, the differences between subgroups, examined by using the post-hoc paired t-tests for $\bm\omega_{4}$ among all the experimental conditions, are reported in Table S-I and Table S-II (supplementary file).
The comparison among different height levels within the same box weight condition shows significant difference only for the upper body. Instead, the comparison among different box weights, especially when considering $10.0$ kg, shows significant differences ($p$ $<0.05$) in most of the cases/joints. 
Due to the large datasets, results of the full statistical analysis are not reported for all the indexes but only for the overloading joint torque index $\bm\omega_{4}$, which was able to provide more relevant information (i.e. the dynamic aspects) for this task. 
However, concerning $\bm\omega_{5}$ and $\bm\omega_{6}$, no significant differences were found for the upper body while in the lower body the results were comparable to $\bm\omega_{4}$, in line with the fact that the former are a function of the latter. Conversely, for $\bm\omega_{7}$ no significant differences were present. 

Considering the outcome of the sEMG investigation, the trend of the muscle activity was comparable to the one of $\bm\omega_{4}$ in most of the considered muscles, indicating its capability to address human physical effort. Specifically, the percentage increase of the muscle activity between the $2.5$ kg box and the $5.0$/$10.0$ kg boxes (averaged among height levels which showed less significant variations) was  $51.3/112.9$ \% in the AD, $33.4/115.9$  \% in the PD, $67.9/175.1$ \% in the BC, $9.1/69.3$ \% in the TC, $49.8/143.5$ \% in the TR and $27.1/90.1$ \% in the ES, $13.2/33.5$ \% in the GM, $23.8/57.6$ \% in the RF, respectively \footnote{The reported values of sEMG are the ones exhibited by the muscles in the same instant in which the maximum value of the compared index $\bm\omega$, which is shown in the polar plots, was found.}. The muscle activity in BF and TA instead was approximately constant among different experimental conditions. The percentage differences of all the dynamic indexes and the associated muscle activity for task $1$ can be found in Table S-III (supplementary file).

\begin{figure}[h]
    \centering
    \includegraphics[width=0.5\textwidth]{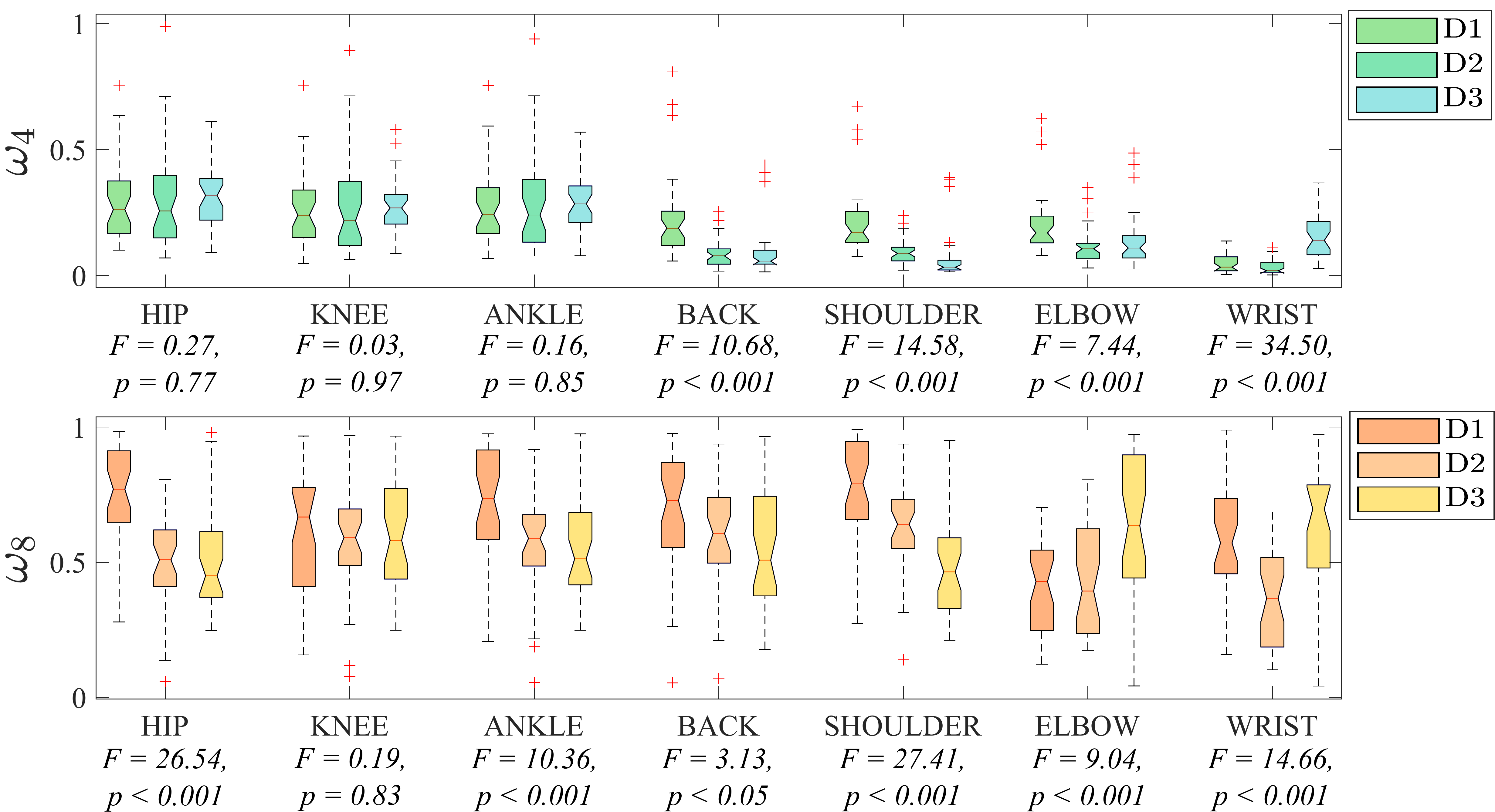}
    \caption{Boxplots for the overloading joint torque index $\bm\omega_{4}$ and the compressive force index $\bm\omega_{8}$ averaged among all the trials and subjects for the experimental conditions of task $2$, i.e. the positions (D1, D2, D3) of the panels to be drilled, in the human main joints. The $F$-ratio ($F$) and $p$-values ($p$) resulting from the ANOVA are also reported.}
    \label{fig:boxplot_2}
\end{figure}

Let us then observe the plots relative to task $2$, illustrated in Figure \ref{fig:Task002_polar}. Considering the indexes accounting for the kinematic aspects (pink area), it can be noted that either $\bm\omega_{1}$, $\bm\omega_{2}$ and $\bm\omega_{3}$ showed low values in all the experimental conditions, except the joint displacement at the elbow level (in line with the task requirements), meaning that kinematic variables are not significantly predictive of the ergonomic risk associated with this specific activity. This is consistent with the fact that the subjects were basically steady throughout its whole duration. On the other hand, considering the indexes accounting for the dynamic aspects, the predominance of $\bm\omega_{8}$ over the other indexes is clearly noticeable. The compressive forces index was overall far higher than all the other indexes and presented a certain increment in D1 with respect to D2 and D3, most likely due to the fact that in D1 the gravitational component of the external force was greater because of the tool weight. In particular, in Figure \ref{fig:boxplot_2}, the boxplots representing the values of the overloading joint torque index $\bm\omega_{4}$ and the compressive force index $\bm\omega_{8}$ averaged among all the trials and subjects for all the experimental conditions of task $2$,
are illustrated in the human main joints.
These two indexes, among the dynamics ones, are expected to provide more relevant information with respect to the type of task. In fact, similarly as for joint kinematics, neither overloading power nor CoM potential energy showed no meaningful variations, again in line with the fact that subjects were steady throughout the task. 
In Figure \ref{fig:boxplot_2}, the $F$-ratio ($F$) and $p$-values ($p$) resulting from the ANOVA are also reported while in Table S-IV (supplementary file), the $p$-values resulting from the post-hoc paired t-tests among all the experimental conditions are presented. As it can be observed in the boxplots, $\bm\omega_{4}$ was quite low in the lower body and even less elevated in the upper body. The differences among drilling locations were significant for the back and the arm joints (p $<.001$), but not for the leg joints. On the other hand, $\bm\omega_{8}$ showed substantially higher values and more variability between joints and experimental conditions. The differences among drilling locations were significant for all the considered joints except the knee. Similarly, the results of the paired t-tests indicate significant differences in more cases/joints for $\bm\omega_{8}$ than for $\bm\omega_{4}$.
As a matter of fact, the overloading joint torque, as it is defined, considered only the gravitational component of the force, i.e. the weight of the tool. Since the tool was partially supported by the drilled interface while drilling, such a component was not significant during this specific task. On the other hand, the compressive force index took into account all the components of the external force induced on the human end-effector since this information was provided by the F/T sensor mounted purposely on the driller. 

Considering the muscles activity analysis, similarly as for $\bm\omega_{8}$, sEMG signals displayed more significant values in D1 than in the other two conditions in several muscles, meaning that $\bm\omega_{8}$ correlated to muscle activity to some extent. Specifically, the percentage increase of the muscle activity between D2/D3 and D1 was $130.3/1153.1$ \% in the AD, $32.2/93.0$ \% in the BC $6.3/267.7$ \% in the TR, $112.7/24.1$  \% in the ES, $3.5/18.9$ \% in the GM, $358.7/106.1$ \% in the RF and $47.6/85.2$ \% in the TA, respectively. The muscle activity in PD was $86.7$ \% lower in D2 but $14.9$ \% higher in D3 than in D1, probably due to the task requirements (i.e. the position of the arm while holding the driller). The same explanation can be given for the muscle activity in TC, which was highest in D3, lower in D2, and even less significant in D1.  The percentage differences of all the dynamic indexes and the associated muscle activity for task $2$ can be found in Table S-V (supplementary file).

\begin{figure}[!h]
    \centering
    \includegraphics[width=0.5\textwidth]{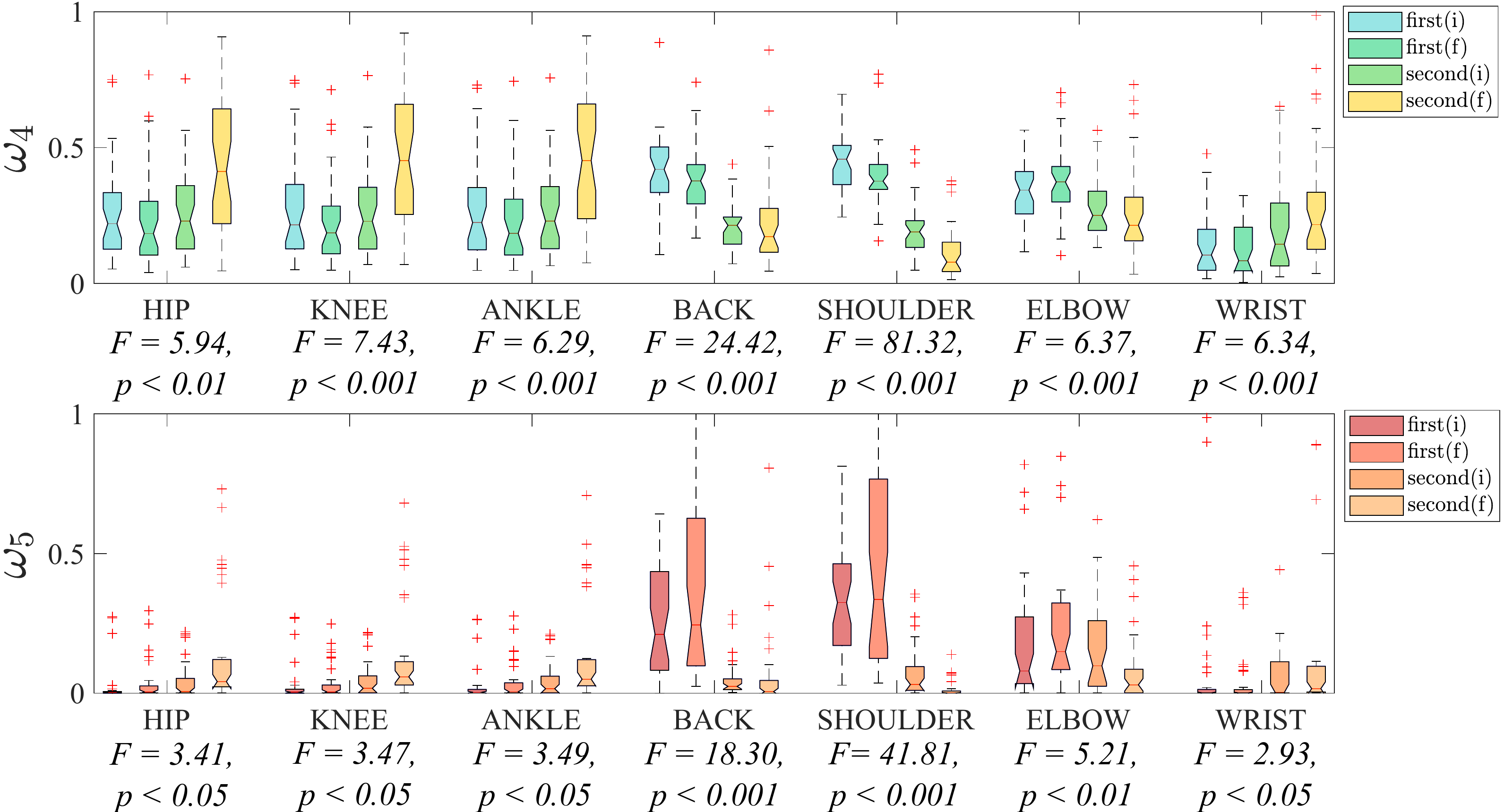}
    \caption{Boxplots for the overloading joint torque index $\bm\omega_{4}$ and the overloading joint fatigue index $\bm\omega_{5}$ averaged among all the trials and subjects for the initial (i) and final (f) instants of the phases (first and second) of task $3$, i.e. one of the two helmets to be painted, in the human main joints. The $F$-ratio ($F$) and $p$-values ($p$) resulting from the ANOVA are also reported.}
    \label{fig:boxplot_3}
\end{figure}

Lastly, let us focus on the plots depicted in Figure \ref{fig:Task003_polar}, relative to task $3$. Considering the kinematics-related variables (pink area), it can be observed that rather meaningful values were shown by $\bm\omega_{1}$ at the elbow and knee level, to a lesser degree at the hip level, meaning that potentially risky body configurations were adopted by the subjects to accomplish the task. On the other hand, $\bm\omega_{2}$ and $\bm\omega_{3}$ were altogether very low thus bearing no information for the purposes of the ergonomics assessment. Similarly as in task $1$, the only exception was the acceleration of the wrist that, as already mentioned, may be suggestive of potential hazards. Considering then the dynamics-related variables, it is evident that the most relevant indexes to explain the physical exposure of the task were $\bm\omega_{4}$ and $\bm\omega_{5}$, in comparison with all the other indexes that are basically negligible. 
In Figure \ref{fig:boxplot_3}, the boxplots representing the values of the overloading joint torque index $\bm\omega_{4}$ and the overloading joint fatigue index $\bm\omega_{5}$ averaged among all the trials and subjects for the initial (i) and final (f) instants of the phases of task $3$, i.e. one of the two helmets to be painted, are illustrated in the human main joints. Concerning the lower body, the overloading torque showed low/moderate values while the overloading fatigue was less considerable. Concerning the upper body, the overloading torque presented higher values in the first phase for all joints except the wrist but its level remained moderate. Similarly, the overloading fatigue exhibited higher values in the first phase but, on the contrary, it reaches a considerable level, mostly in the shoulder and in the back, as it can be noted from the considerable standard deviation values. In fact, as already explained, $\bm\omega_{4}$ considered the instantaneous effect of the external load while $\bm\omega_{5}$ was capable to address its cumulative effect thus accounting for the accumulation of fatigue. Moreover, $\bm\omega_{5}$ presented more variability between different trials and subjects since fatigue occurred differently among individuals as explained by the subject-specific parameters that the model is based on. Nevertheless, what is not evident observing these results but it should be underlined, is that the overloading fatigue model was extremely sensitive to the threshold which has been set to distinguish among fatigue and recovery phase. In Figure \ref{fig:boxplot_3}, the $F$-ratio ($F$) and $p$-values ($p$) resulting from the ANOVA are also reported. The differences among the considered instants were significant for all the considered joints for both $\bm\omega_{4}$ and $\bm\omega_{5}$. On the other hand, in Table S-VI (supplementary file), the $p$-values resulting from the post-hoc paired t-tests among all the experimental conditions are reported. Significant differences can be found mostly in the back and in the shoulder for both the considered indexes, in line with the fact that these are more involved joints in task $3$.  
In addition, the trend of the muscle activity was comparable to the one displayed by both the overloading torque and fatigue in the lower body and in the arm, suggesting the capability of $\bm\omega_{4}$ and also $\bm\omega_{5}$ to account for human effort. Specifically, the raising of the overloading torque and fatigue in the legs between the first and the second phase was accompanied by a percentage increase of muscle activity of $348.5$ \% in the RF, $177.5$ \% in the BF, and $624.5$ \% in the TA. Conversely, the overloading torque and fatigue in the arms were reduced between the first and the second phase but still matching the activity in most of the muscles with a percentage decrement of $78.2$ \% in the AD, $15.2$ \% in the BC, $48.8$ \% in the TC and $42.9$ \% in the TR.  The percentage differences of all the dynamic indexes and the associated muscle activity for task $3$ can be found in Table S-VII (supplementary file). 

\subsubsection{Analysis with EAWS method}

As mentioned above, to assess the potential of the proposed set of ergonomic indexes, its outcome was compared with the ergonomic risk scores provided by a well-recognised and widely used tool to evaluate workers' physical load, namely, the EAWS method \cite{schaub2013european}.
It is critical to note that EAWS were conceived and usually applied in real industrial scenarios. On the contrary, for the sake of this experimental analysis, they were adopted with simplified activities that were designed to simulate occupational tasks and were conducted during experiments in laboratory settings. Accordingly, several assumptions were made to contextualise such simplified tasks in the workplace that was intended to be replicated.
Since the level of performance of the recruited subjects was not realistic compared to an actual worker, the task timing was defined according to standard base time using the Methods-Time Measurement - Universal Analyzing System (MTM-UAS) method \cite{maynard1948methods}. In fact, in the industrial environment, it is necessary to quantify work with respect to a regulatory level of performance \cite{caragnano2018ergo}.
The EAWS method enables a comprehensive and unique ergonomic risk evaluation. In fact, four different sections that focus on each specific aspect of manual material activities are included and their results can be integrated into a combined score presented in an intuitive traffic light scheme (green, yellow, red) according to the Machinery Directive 2006/42/EC (EN 614). Accordingly, the evaluation of the three occupational tasks considered in this experimental analysis by means of the EAWS method resulted in a total score for each of them, indicative of the associated ergonomic risk level. 

The EAWS score was computed for each trial of every task and then the mean value among the resulting score was computed. This procedure was repeated for all twelve subjects. As a result, each subject presented an EAWS final score for each task. It should be noted that for task $1$ three different ratings were computed, considering the box weight conditions separately. The mean and the standard deviation values which were finally computed among all the subjects and the corresponding risk level are illustrated in Table \ref{tab:EAWSscore}.
\begin{table}[h!]
	\centering
	\caption{Mean and standard deviation\footnoteref{footnote_eaws} computed among twelve subjects of the EAWS final scores for all the task considered in the experimental analysis.}
	\label{tab:EAWSscore}
	\begin{adjustbox}{max width=0.48\textwidth}
	\begin{tabular}{ccccc}
	    \toprule
	    \textbf{Task} & \textbf{Activity} & \textbf{Condition} & \textbf{EAWS Final Score} & \textbf{Risk level\footnotemark} \\
	    \midrule
	    1 & lifting/lowering & $2.5$ kg box & $38.8$ $\pm$ $1.2^{\dag\dag}$ & Moderate \\
	     & & $5.0$ kg box & $41.7$ $\pm$ $1.7^{\dag\dag}$ & Moderate\\
	     & & $10.0$ kg box & $48.7$ $\pm$ $6.9^{\dag\dag}$ & Moderate \\
	    \midrule
	    2 & drilling & & $23.1$ $\pm$ $5.0^{\dag}$ & Low\\
	    \midrule
	    3 & painting & & $14$ $\pm$ $0.0^{\dag}$ & Low \\
	    \bottomrule
	\end{tabular}
	\end{adjustbox}
\end{table}
\footnotelabeled{footnote_eaws}{The risk level associated with the EAWS score was determined as follows: the range $0-25$ points corresponded to a low risk ($^{\dag}$), the range $25-50$ points corresponded to a moderate risk ($^{\dag\dag}$) and a value $>50$ points corresponded to a high risk, respectively.}

For task $1$, the inter-subject variability of the EAWS score only depended on the worker gender, i.e. the male and the female subjects, respectively, had the exact same rating among them. This was due to the level of risk associated with the box weight which was indeed gender-specific. It can be noted that, as expected, the EAWS score of the lifting/lowering task grown with increasing box weight. Looking at the mean among the subjects, the risk level associated was moderate ($^{\dag\dag}$) for all the experimental conditions. Nevertheless, considering the $10.0$ kg box one, the EAWS score was quite close to the higher threshold ($>50$) and the standard deviation equal to $6.9$ suggests that for some subjects the risk level turned into high. It should be observed that the growing trend of the risk level with the box weight was likewise identified by the proposed human ergonomics monitoring framework.
For task $2$, each trial of each subject exhibited a rather different EAWS score since the maximum value (the worst condition) of the force exerted while drilling was taken into account. Similarly, as for the last experimental condition in task $1$, the rating was quite close to the mid threshold ($>25$) and presented a significant standard deviation ($5.0$). Thus, the risk level considering the mean value was low ($^{\dag}$) but for some subjects it turned into moderate. 
On the other hand, for task $3$, the EAWS score was exactly the same for each subject, leading to a standard deviation equal to $0.0$.  In fact, neither gender nor other individual factors influenced the evaluation of this specific activity but only the task timing. The risk level associated was then low for all the subjects, abundantly within the low part of the range.

\subsection{Discussion}

In view of the results presented in the previous paragraph, some remarks can be drawn for the human ergonomics monitoring framework proposed in this study. 
As regards kinematics-related variables, the most beneficial index in assessing physical load for the considered task was represented by $\bm\omega_{1}$. The joint displacement index allows to detect wherever the human current body configuration lays within specific sections of the human RoM which should be avoided (e.g. in proximity to the maximum limits). Hence, by monitoring how often and how long these body configurations are maintained, it is possible to determine if a potential ergonomic risk exists. As a matter of fact, joint kinematics, in general, has extensively been used in the analysis of human movement and joint angles, in particular, have widely been employed to identify human awkward or unfavourable postures while performing working activities. Concerning joint velocities and accelerations, no significant information was offered by $\bm\omega_{2}$ and $\bm\omega_{3}$ in the context of this experimental analysis, excluding the rather significant accelerations exhibited by the wrist, which should be taken into account since they were demonstrated to correlate with occupational risks. Nevertheless, the three tasks considered were quite steady and regular and did not require the development of significant speeds and accelerations. Hence, the joint velocity and acceleration indexes should be investigated more thoroughly.

Concerning dynamics-related variables, for each considered task, the ergonomic indexes that better explain the required physical load were identified, as evaluated through statistical analysis and supported by the outcome of a sEMG investigation. Specifically, the overloading joint torque index $\bm\omega_{4}$ was the more promising mean to account for the mechanical overburden of the body structures in tasks involving the handling of heavy objects (task $1$). It showed far higher values than the other indexes with significant differences among all the experimental conditions and it correlated with the muscle activity. Instead, considering a task in which all the components of the interaction forces at the hand/tool interface are relevant - not only the gravitational one - (task $2$), the joint compressive forces index $\bm\omega_{8}$ proved to be more predictive of the risk associated with dynamically varying interaction forces. In fact, in these cases, the overloading joint torque may underestimate the hazard. Finally, concerning repetitive and monotonous prolonged activities (task $3$) both the overloading joint torque and fatigue indexes, $\bm\omega_{4}$ and $\bm\omega_{5}$, respectively, exhibited significant results. The overloading fatigue index happened to be more valuable since it better accounts for the variability among the subjects. Moreover, it took into consideration the cumulative effect of the risk and not the instantaneous one thus it was able to account for the accumulation of fatigue. However, the sensitivity of the overloading fatigue model to thresholds must be further investigated, also considering longer-lasting activities. Similarly, concerning the overloading joint power index $\bm\omega_{6}$ and the CoM potential energy index $\bm\omega_{7}$ a deeper examination should be made. Not remarkable or not easily interpretable information appeared to be provided by such variables but this may be due to the fact that, as already noticed, no considerable velocities or unexpected movements were required to accomplish the analysed tasks. Conversely, regular, smooth, and constant-speed actions were repeated. Hence, the potential of both the indexes should be investigated more deeply while conducting more dynamically varying jobs, likewise for $\bm\omega_{2}$ and $\bm\omega_{3}$.

Ultimately, a comparison should be made between the proposed set of ergonomic indexes and the EAWS. Since the EAWS only provides a single score for each task rather than a trend, the similarities of the differences among the experimental conditions were simply analysed. 
It should be observed that EAWS happened to be subject-specific only to a certain degree. In fact, the inter-subject variations were significant only for task $2$. For task $1$, only gender was crucial to the computation of the final score
while for task 3, the exact same score was assigned to all the subjects. 
Consequently, it is possible to state that the set of ergonomic indexes proposed in this study pays much more attention to the individual's behaviours and demands due to a subject-specific model of the human body.
In addition, the EAWS analysis consists in a rather complex and articulated procedure that can be conducted only in an off-line phase and, according to the author, to some extent is affected by the subjective opinions of the expert who carries out the analysis. 
Finally, this method seems still to lack a full evaluation of all the dynamic aspects underlying the execution of occupational tasks. 

\section{Human Measurements Database}
\label{sec:database}

The created database including the data from all the sensors system employed is available in Zenodo\footnote{Digital Object Identifier: 10.5281/zenodo.5575139} and is organised as follows. There is a folder for each task (task $1$: lifting/lowering of a heavy object, task $2$: drilling, task $3$: painting with a lightweight tool). Inside, there is a folder for each one of the twelve participants (named after the subject’s ID) that includes a file containing the ID, age, gender, handedness, body mass, body dimensions, MVC values of the subject, and then, four folders containing the data in the available formats, i.e. \textit{.mat}, \textit{.bag}, \textit{.mp4} and \textit{.mvnx} (plus \textit{.c3d}). In all these folders, there is a file for each one of the trials performed by the subjects. 
Each \textit{.mat} is a binary MATLAB\textregistered~ (MathWorks, Inc.; headquarters: Natick, MA, United States) file that stores workspace variables and contains the information that is listed in Table \ref{tab:database}, divided according to the sensor system employed to collect it. Similarly, each \textit{.bag} file contains the same information but in the form of ``ROS bags'' that are a file format developed in ROS for storing ROS message data and can be conveniently played back (details can be found in \cite{ROS}). Each \textit{.mp4} file contains the video recording of each trial to facilitate the data usage. Finally, each \textit{.mvnx} is the Xsens custom open XML format that contains all the information collected by the Xsens system. Also the \textit{.c3d} exported from the latter is available. Besides the participants' folders, in the database home folder, some additional files are also provided to easily manage the data. The latter include: a short guide with general information about the database, a file describing the experimental setup, and a file describing in detail the human model employed and the reference frame adopted.

\begin{table}[]
\centering
\caption{Information included in the human measurements database on factory-like activities divided according to the sensor system employed to collect it.}
	\label{tab:database}
	\begin{adjustbox}{max width=0.5\textwidth}
\begin{tabular}{cl} \toprule
\textbf{Sensor system} & \textbf{Information} \\ \midrule
\multirow{2}{*}{Motion-capture suit} & \begin{tabular}[c]{@{}l@{}} 3D position, orientation (quaternion), linear and angular \\ velocity, and linear and angular acceleration of the origin \\ of the 21 body segments of the human model.\end{tabular} \\ 
 &  Angles of the 20 joints (3 DoFs each) of the human  model.\\ \midrule
\multirow{3}{*}{Force plate} & 2D position of the CoP. \\ 
 & GRF. \\ 
 & Ground reaction moment (GRM). \\ \midrule
sEMG system & Muscle activity of the ten recorded muscles \\ \midrule
\multirow{2}{*}{F/T sensor} & 3D force \\
 & 3D torque \\ \bottomrule
\end{tabular}
\end{adjustbox}
\vspace{-4 mm}
\end{table}

\section{Conclusions}
\label{sec:conclusion}

In this paper, a novel method to assess human ergonomics in the workplace was introduced. Several physical variables are estimated that can lay the foundations for the ergonomic risk categorisation. 
Since the proposed set of indexes was conceived to address multiple physical risk factors, it was capable to evaluate different working activities. Indeed, for each considered task, a subset of ergonomic indexes was found to be predictive of the required physical load. Significant differences were found among different experimental conditions, as evaluated by the statistical analysis and further evidence is provided by the trend of the sEMG signals. The potential of the remaining indexes must be investigated more deeply in future studies considering more dynamically varying and activities. 

The comparison between the EAWS and the proposed set of ergonomic indexes highlighted the pros and cons of the two approaches. Hence, upcoming studies may focus on their integration, allowing a more comprehensive and thorough analysis of the occupational risk factors, taking equally into account the relative kinematics and dynamic aspects and addressing workers' specific requirements.

The proposed framework was developed envisaging its use in an industrial environment thus it can operate with fit-for-workplaces (i.e. wearable, long-life, and easy-to-use) sensor systems. In this work, the computations were made off-line to ensure higher accuracy but in future studies its online capabilities will be demonstrated. 

A force plate was employed for the sake of a thorough validation, however, it can be replaced with sensor insoles. On the other hand, a wearable inertial-based system was employed to track human motion but vision-based systems can be considered as an alternative.

A sagittal model was employed to represent the human body since only activities performed along the sagittal plane were analysed. Nevertheless, its extension to a 3D model to consider more complex tasks will be investigated in the future.

Lastly, the database created in this study would serve a twofold purpose. First, to advance the research in the monitoring and assessment of human ergonomics in the context of the ERC project Ergo-Lean, that aims to propose collaborative robotics technology to improve human physical factors in the workplace. Next, to share experimental data with the scientific community for comparisons in ergonomics studies as well as model identification or learning for assistive and collaborative robotics. In future works, the extension and enrichment of this database will be covered as well.

\ifCLASSOPTIONcaptionsoff
  \newpage
\fi

\bibliographystyle{IEEEtran}
\bibliography{biblio}

\end{document}